\documentclass[journal]{IEEEtran}
\usepackage{amssymb}
\usepackage{amsmath}
\usepackage{amsfonts}
\usepackage{cite}
\usepackage{graphics}
\usepackage{epsf}
\usepackage{float}
\usepackage[pdftex]{graphicx}
\usepackage[small]{caption2}
\usepackage{bm}
\usepackage{acronym}
\usepackage{multirow}
\usepackage{color}
\usepackage[normalem]{ulem}
\usepackage{array}

\newcommand{\be}{\begin{equation}}
\newcommand{\ee}{\end{equation}}
\newcommand{\bea}{\begin{eqnarray}}
\newcommand{\eea}{\end{eqnarray}}
\newcommand{\beaa}{\begin{eqnarray*}}
\newcommand{\eeaa}{\end{eqnarray*}}

\acrodef{1D}[1D]{one-dimensional}
\acrodef{2D}[2D]{two-dimensional}
\acrodef{GTBWT}[GTBWT]{generalized tree-based wavelet transform}
\acrodef{RTBWT}[RTBWT]{redundant tree-based wavelet transform}

\title{Image Processing using Smooth Ordering of its Patches}

\begin{document}
\sloppy

\author{Idan Ram, Michael~Elad,~\IEEEmembership{Fellow,~IEEE}, and Israel~Cohen,~\IEEEmembership{Senior~Member,~IEEE}
\thanks{
I. Ram and I. Cohen are with the Department of Electrical Engineering, Technion --
Israel Institute of Technology, Technion City, Haifa 32000, Israel. E-mail
addresses: idanram@tx.technion.ac.il (I. Ram), icohen@ee.technion.ac.il
(I. Cohen); tel.: +972-4-8294731; fax: +972-4-8295757.
M. Elad is with the Department of Computer Science, Technion --
Israel Institute of Technology, Technion City, Haifa 32000, Israel. E-mail
address: elad@cs.technion.ac.il
} }

\maketitle

\begin{abstract}

We propose an image processing scheme based on reordering of its patches.
For a given corrupted image, we extract all patches with overlaps,
refer to these as coordinates in high-dimensional space,
and order them such that they are chained in the "shortest possible path",
essentially solving the traveling salesman problem.
The obtained ordering applied to the corrupted image, implies a permutation of the image pixels to {\em what should be} a regular signal.
This enables us to obtain good recovery of the clean image
by applying relatively simple \ac{1D} smoothing operations (such as filtering or interpolation) to the reordered set of pixels.
We explore the use of the proposed approach to image denoising and inpainting,
and show promising results in both cases.

\end{abstract}


\section{Introduction}

In recent years, image processing using local patches has become very popular and was shown to be highly effective -- see [1-11] for representative work.
The core idea behind these and many other contributions is the same: given the image to be processed,
extract all possible patches with overlaps; these patches are typically very small compared to the original image size (a typical patch size would be $8 \times 8$ pixels).
The processing itself proceeds by operating on these patches and exploiting interrelations between them.
The manipulated patches (or sometimes only their center pixels) are then put back into the image canvas to form the resulting image.

There are various ways in which the relations between patches can be taken into account:
weighted averaging of pixels with similar surrounding patches, as the NL-Means algorithm does \cite{buades2006review},
clustering the patches into disjoint sets and treating each set differently, as performed in
\cite{chatterjee2009clustering},\cite{yu2010image},\cite{yu2010solving}\cite{dong2011sparsity},\cite{zoran2011learning},
seeking a representative dictionary for the patches and using it to sparsely represent them, as practiced in \cite{elad2006image},\cite{mairal2008sparse},\cite{mairal2009non} and \cite{zeyde2012single},
gathering groups of similar patches and applying a sparsifying transform on them \cite{mairal2009non},\cite{dabov2007image}.
A common theme to many of these methods is the expectation that every patch taken from the image may find similar ones extracted elsewhere in the image.
Put more broadly, the image patches are believed to exhibit a highly-structured geometrical form in the embedding space they reside in.
A joint treatment of these patches supports the reconstruction process by introducing a non-local force, thus enabling better recovery.

In our previous work \cite{ram2011generalized} and \cite{ram2012redundant} we proposed yet another patch-based image processing approach.
We constructed an image-adaptive wavelet transform which is tailored to sparsely represent the given image.
We used a plain 1D wavelet transform and adapted it to the image by operating on a permuted order of the image pixels\footnote{Note that the idea of adapting a wavelet transform to the image by reordering its pixels appeared already in \cite{plonka2009easy}, but the scheme proposed there did not use image patches, and targeted image
compression only.}.
The permutation we proposed is drawn from a shortest path ordering of the image patches.
This way, the patches are leveraged to form a multi-scale sparsifying global transform for the image in question.

In this paper we embark from our earlier work as reported in \cite{ram2011generalized} and \cite{ram2012redundant},
adopting the core idea of ordering the patches.
However, we discard the globality of the obtained transform and the processing that follows, the multi-scale treatment, and the sparsity-driven processing that follows.
Thus, we propose a very simple image processing scheme that relies solely on patch reordering.
We start by extracting all the patches of size $\sqrt{n}\times\sqrt{n}$ with maximal overlaps.
Once these patches are extracted, we disregard their spatial relationships altogether, and seek a {\em new way} for organizing them.
We propose to refer to these patches as a cloud of vectors/points in $\mathbb{R}^n$,
and we order them such that they are chained in the "shortest possible path",
essentially solving the traveling salesman problem \cite{cormen2001introduction}.
This reordering is the one we have used in \cite{ram2011generalized} and \cite{ram2012redundant}, but as opposed to our past work,  our treatment from this point varies substantially.
A key assumption in this work is that proximity between two image patches implies proximity between their center pixels.
Therefore if the image mentioned above is of high-quality, the new ordering of the patches is expected to induce a highly regular (smooth or at least a piece-wise smooth)
1D ordering of the image pixels, being the center of these patches.
When the image is deteriorated (noisy, containing missing pixels, etc.), the above ordering is expected to be robust to the distortions, thereby suggesting a reordering of the corrupted pixels to "what should be" a regular signal.
Thus, applying relatively simple \ac{1D} smoothing operations (such as filtering or interpolation) to the reordered set of pixels should enable good recovery of the clean image.

This is the core process we propose in this paper –-- for a given corrupted image, we reorder its pixels, operate on the new 1D signal using simplified algorithms, and reposition the resulting values to their original location.
We show that the proposed method, applied with several randomly constructed orderings and combined with a proposed subimage averaging scheme,
is able to lead to state-of-the-art results.
We explore the use of the proposed image reconstruction scheme to image denoising,
and show that it achieves results similar to the ones obtained with the K-SVD algorithm \cite{elad2006image}.
We also explore the use of the proposed image processing scheme to image inpainting,
and show that it leads to better results compared to the ones obtained with a simple interpolation scheme
and the method proposed in \cite{elad2010sparse} which employs sparse representation modeling via the redundant DCT dictionary.
Finally, we draw some interesting ties between this scheme and BM3D rationale \cite{dabov2007image}.

The paper is organized as follows:
In Section II we introduce the basic image processing scheme.
In Section III we explain how the performance of the basic scheme can be improved using a subimage averaging scheme,
and describe the connection between the improved scheme and the BM3D algorithm.
In Section IV we explore the use of the proposed approach to image denoising and inpainting,
and present experimental results that demonstrate the advantages of the proposed scheme.
We summarize the paper in Section IV with ideas for future work along the path presented here.

\section{Image Processing using Patch Ordering}

\subsection{The Basic Scheme}
\label{scheme}
Let $\mathbf{Y}$ be an image of size $N_1\times N_2$ where $N_1 N_2=N$,
and let $\mathbf{Z}$ be a corrupted version of $\mathbf{Y}$,
which may be noisy or contain missing pixels.
Also, let $\mathbf{z}$ and $\mathbf{y}$ be the column stacked representations of $\mathbf{Z}$ and $\mathbf{Y}$,
respectively.
Then we assume that the corrupted image satisfies
\begin{equation}
\mathbf{z}=\mathbf{My}+\mathbf{v}
\end{equation}
where the $N\times N$ matrix $\mathbf{M}$ denotes a linear operator which corrupts the data,
and $\mathbf{v}$ denotes an additive white Gaussian noise independent of $\mathbf{y}$ with zero mean and
variance $\sigma^2$.
In this work the matrix $\mathbf{M}$ is restricted to represent a point-wise operator,
covering applications such as denoising and inpainting.
The reason for this restriction is the fact that we will be permuting the pixels in the image,
and thus spatial operations become far more complex to handle.

\begin{figure}[t]
\centering
\includegraphics[scale=0.35]{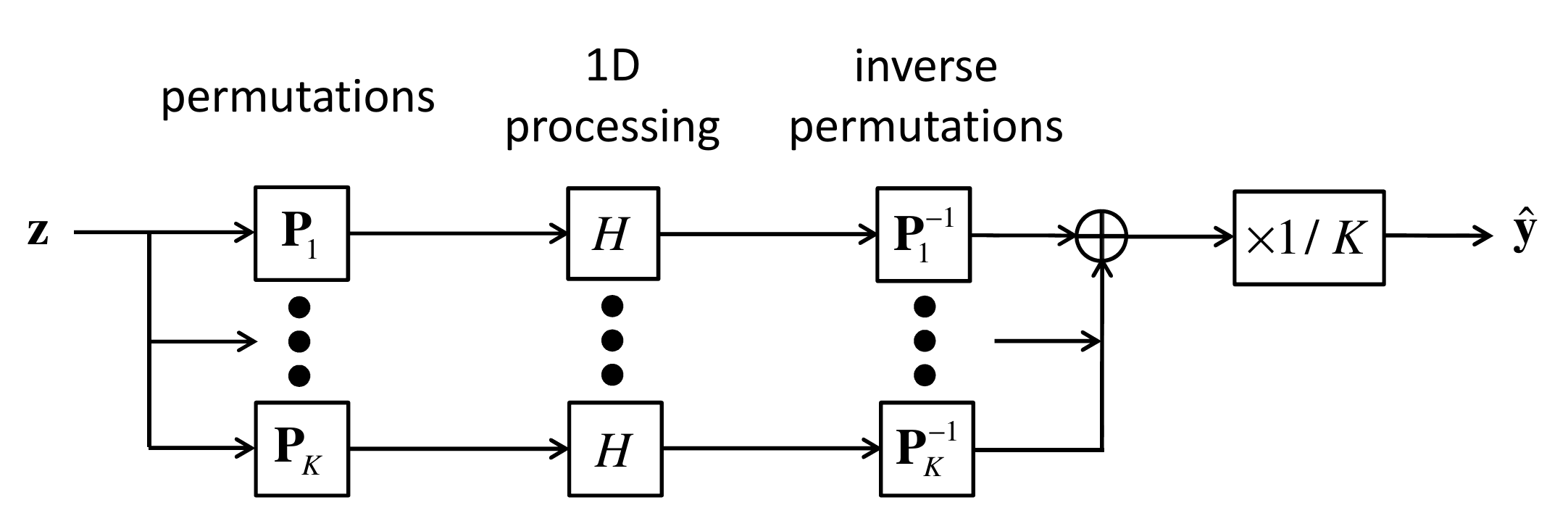}
\caption{The basic image processing scheme.}
\label{Figure: image processing scheme}
\end{figure}

Our goal is to reconstruct $\mathbf{y}$ from $\mathbf{z}$,
and for this end we employ a permutation matrix $\mathbf{P}$ of size $N\times N$.
We assume that when $\mathbf{P}$ is applied to the target signal $\mathbf{y}$,
it produces a smooth signal $\mathbf{y}^p=P\mathbf{y}$.
We will explain how such a matrix may be obtained using the image patches in Section \ref{smooth P}.
We start by applying $\mathbf{P}$ to $\mathbf{z}$ and obtain $\mathbf{z}^p=\mathbf{Pz}$.
Next, we take advantage of our prior knowledge that $\mathbf{y}^p$ should be smooth,
and apply a ``simple'' \ac{1D} smoothing operator $H$ on $\mathbf{z}^p$, such as \ac{1D} interpolation or filtering.
Finally, we apply $\mathbf{P}^{-1}$ to the result, and obtain the reconstructed image
\begin{align}
\hat{\mathbf{y}}=\mathbf{P}^{-1}H\left\{\mathbf{P}\mathbf{z}\right\}.
\end{align}
In order to better smooth the recovered image, we use an approach which resembles the "cycle spinning" method \cite{coifman1995translation}.
We randomly construct $K$ different permutation matrices $\mathbf{P}_k$,
utilize each to denoise the image $\mathbf{z}$ using the scheme described above, and average the results.
This can be expressed by
\begin{align}
\hat{\mathbf{y}}=\frac{1}{K}\sum_{k=1}^{K}\mathbf{P}_k^{-1}H\left\{\mathbf{P}_k\mathbf{z}\right\}.
\end{align}
Fig. \ref{Figure: image processing scheme} shows the proposed image processing scheme.
We next describe how we construct the reordering matrix $\mathbf{P}$.

\subsection{Building the Permutation Matrix $\mathbf{P}$}
\label{smooth P}
We wish to design a matrix $\mathbf{P}$ which produces a smooth signal when it is applied to the target image $\mathbf{y}$.
When the image $\mathbf{Y}$ is known, the optimal solution would be to reorder it as a vector, and then apply a simple {\em sort} operation on the obtained vector.
However, we are interested in the case where we only have the corrupted image $\mathbf{Z}$.
Therefore, we seek a suboptimal ordering operation, using patches from this image.

Let $y_i$ and $z_i$ denote the $i$th samples in the vectors $\mathbf{y}$ and $\mathbf{z}$, respectively.
We denote by $\mathbf{x}_i$ the column stacked version of the $\sqrt{n}\times\sqrt{n}$ patch around the location of $z_i$ in $\mathbf{Z}$.
We assume that under a distance measure \footnote{Throughout this paper we will be using variants of the squared Euclidean distance.} $w(\mathbf{x}_i,\mathbf{x}_j)$,
proximity between the two patches $\mathbf{x}_i$ and $\mathbf{x}_j$ suggests proximity between the uncorrupted versions of their center pixels $y_i$ and $y_j$.
Thus, we shall try to reorder the points $\mathbf{x}_i$ so that they form a smooth path,
hoping that the corresponding reordered \ac{1D} signal $\mathbf{y}^p$ will also become smooth.
The ``smoothness'' of the reordered signal $\mathbf{y}^p$ can be measured using its total-variation measure
\begin{equation}
\|\mathbf{y}^p\|_{TV}=\sum_{j=2}^{N} |y^p(j)-y^p(j-1)|.
\end{equation}
Let $\{\mathbf{x}_j^p\}_{j=1}^N$ denote the points $\{\mathbf{x}_i\}_{i=1}^N$ in their new order.
Then by analogy, we measure the "smoothness" of the path through the points $\mathbf{x}_j^p$ by the measure
\begin{equation}
X_{TV}^p=\sum_{j=2}^{N} w(\mathbf{x}_j^p,\mathbf{x}_{j-1}^p).
\end{equation}
Minimizing $X_{TV}^p$ comes down to finding the shortest path that passes through the set of points $\mathbf{x}_i$, visiting each point only once.
This can be regarded as an instance of the traveling salesman problem \cite{cormen2001introduction},
which can become very computationally expensive for large sets of points.
We choose a simple approximate solution,
which is to start from a random point and then continue from each point $\mathbf{x}_{j_0}$
to its nearest neighbor $\mathbf{x}_{j_1}$ with a probability $p_1\propto \exp\left[-\frac{w(\mathbf{x}_{j_0},\mathbf{x}_{j_1})}{\epsilon}\right]$,
or to its second nearest neighbor $\mathbf{x}_{j_2}$ with a probability $p_2\propto \exp\left[-\frac{w(\mathbf{x}_{j_0},\mathbf{x}_{j_2})}{\epsilon}\right]$,
where $\epsilon$ is a design parameter, and $\mathbf{x}_{j_1}$ and $\mathbf{x}_{j_2}$ are taken from the set of unvisited points.

We restrict the nearest neighbor search performed for each patch to a surrounding square neighborhood which contains $B\times B$ patches.
When no unvisited patches remain in that neighborhood, we search for the nearest neighbor among all the unvisited patches in the image.
This restriction decreases the overall computational complexity, and our experiments show that with a proper choice of $B$ it also leads to improved results.
The permutation applied by the matrix $\mathbf{P}$ is defined as the order in the found path.

\begin{figure}[t]
\centering
\begin{tabular}{cc}
\includegraphics[scale=0.2]{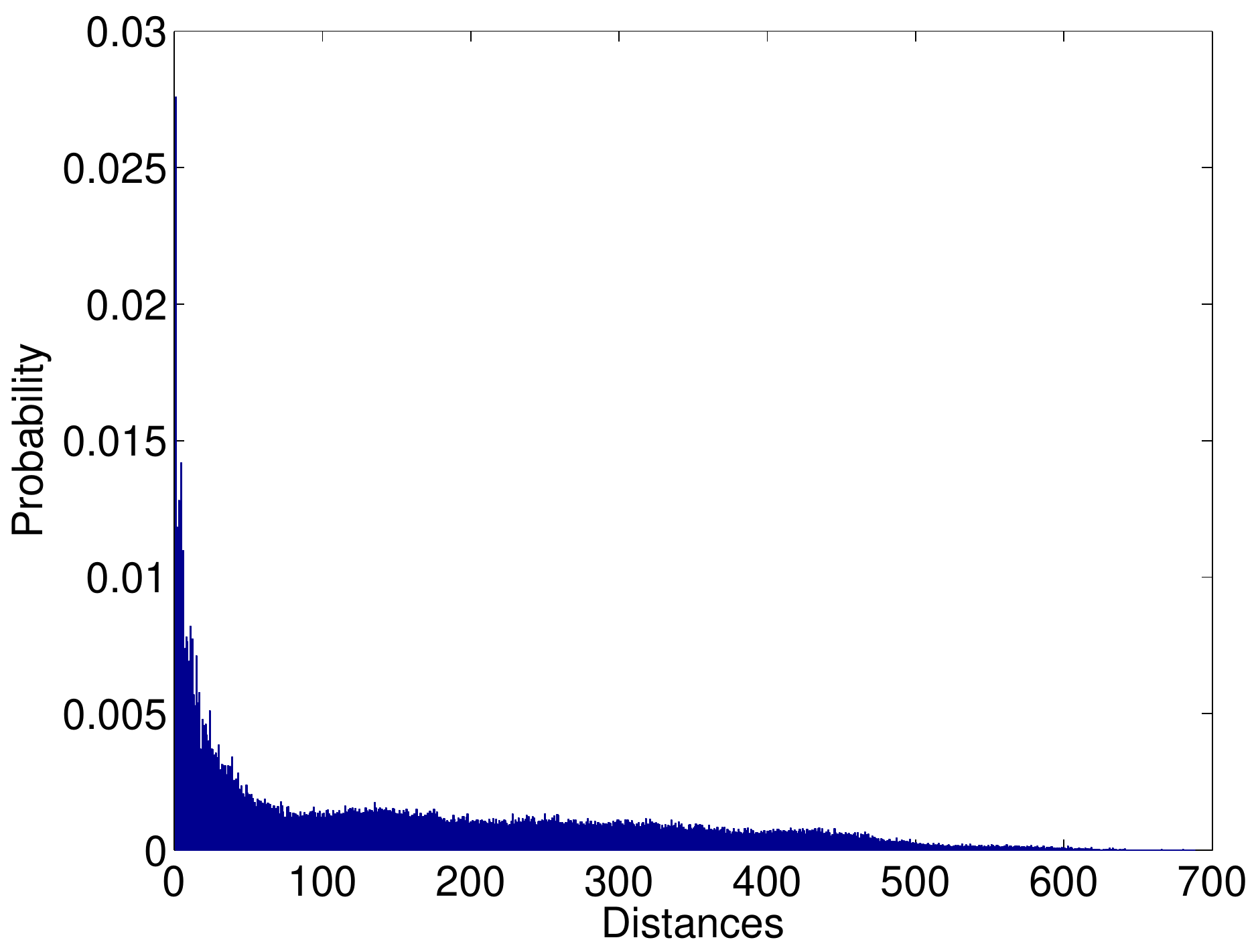} & \includegraphics[scale=0.2]{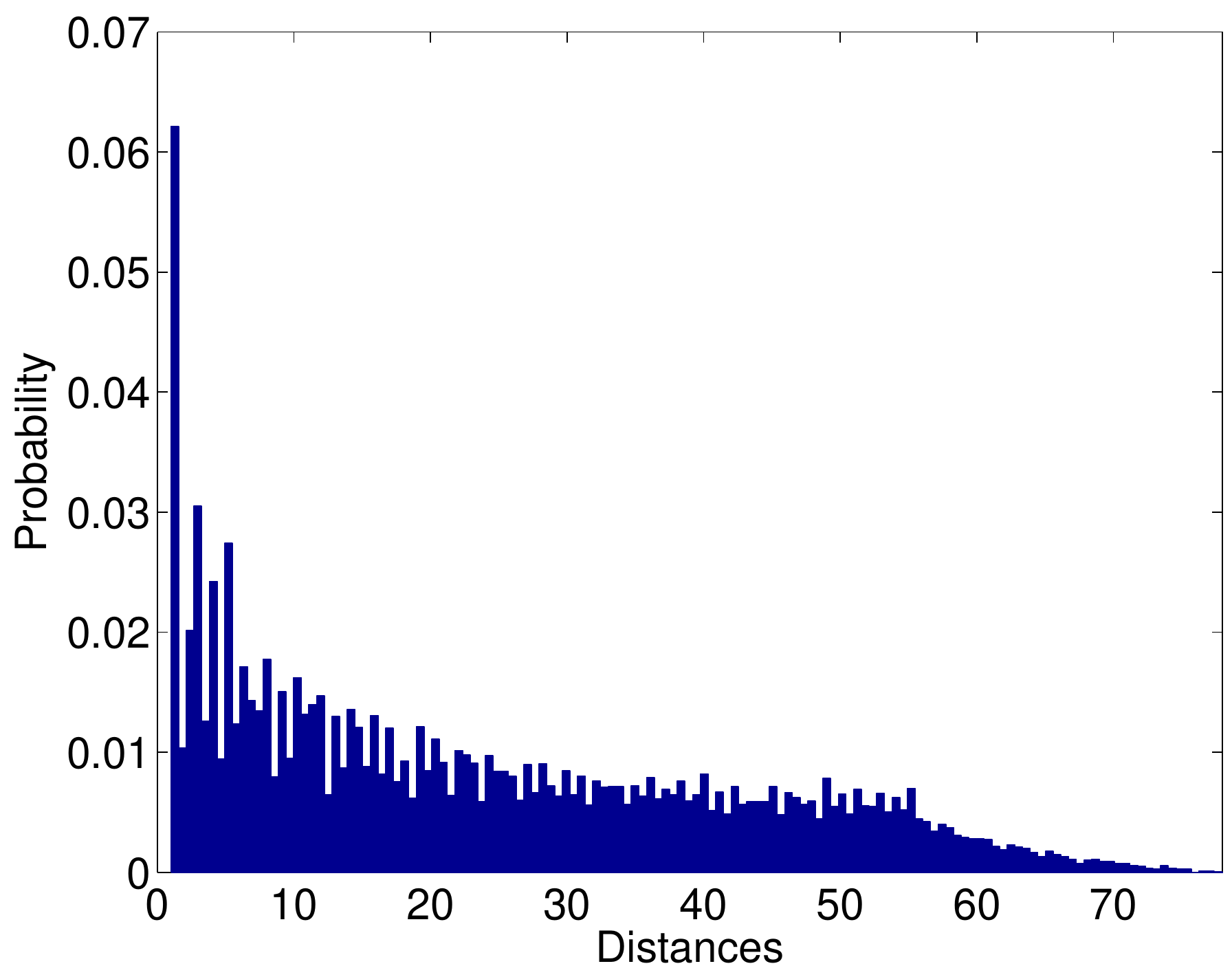}\\
{\small (a)} & {\small (b)}\\
\end{tabular}
\caption{Two normalized histograms of the spatial distances between adjacent patches after the reordering, obtained for a noisy version of the image Barbara with unrestricted (a) and restricted (b) search areas.
The lengths of the pathes in the spatial domain obtained with the unrestricted and restricted searches were
$4.2\cdot10^7$ and $6.1\cdot10^6$, respectively.}
\label{Figure: dist hists}
\end{figure}

It is interesting to examine the characteristics of the patch ordering in the spatial domain.
To this end we apply the patch ordering scheme described above to the patches of the noisy Barbara image shown in Fig. \ref{Figure: subimages}(a) with both unrestricted and restricted ($B=111$) search neighborhoods, and with the parameters $\sqrt{n}=8$ and $\epsilon=10^6$.
We apply the obtained permutations to the patches,
and calculate two normalized histograms of the spatial distances between adjacent patches,
shown in Figure \ref{Figure: dist hists}.
Fig. \ref{Figure: dist hists}(a) shows that only about $3\%$ of neighboring patches in the path are also immediate spatial neighbors,and that far away patches are often assigned as neighbors in the reordering process.
The histogram in Fig. \ref{Figure: dist hists}(b) is limited to show only distances which are smaller or equal to 78,
the maximal possible distance within the search window.
It can be seen that despite the restriction to a smaller search neighborhood,
only about $6\%$ of neighboring patches in the path are also immediate spatial neighbors,
and patches all over the search neighborhood are assigned as neighbors in the reordering process.

In order to facilitate the cycle-spinning method mentioned above, we simply run the proposed  ordering solver $K$ times,
and the randomness (both in the initialization and in assigning the neighbors) leads to different permutation results.
We next describe how the quality of the produced images may be further improved using a subimage averaging scheme,
which can be seen as another variation of "cycle spinning".

\subsection{Subimage Averaging}
\label{subimage_averaging}
\begin{figure}[t]
\centering
\begin{tabular}{cc}
\includegraphics[scale=0.4]{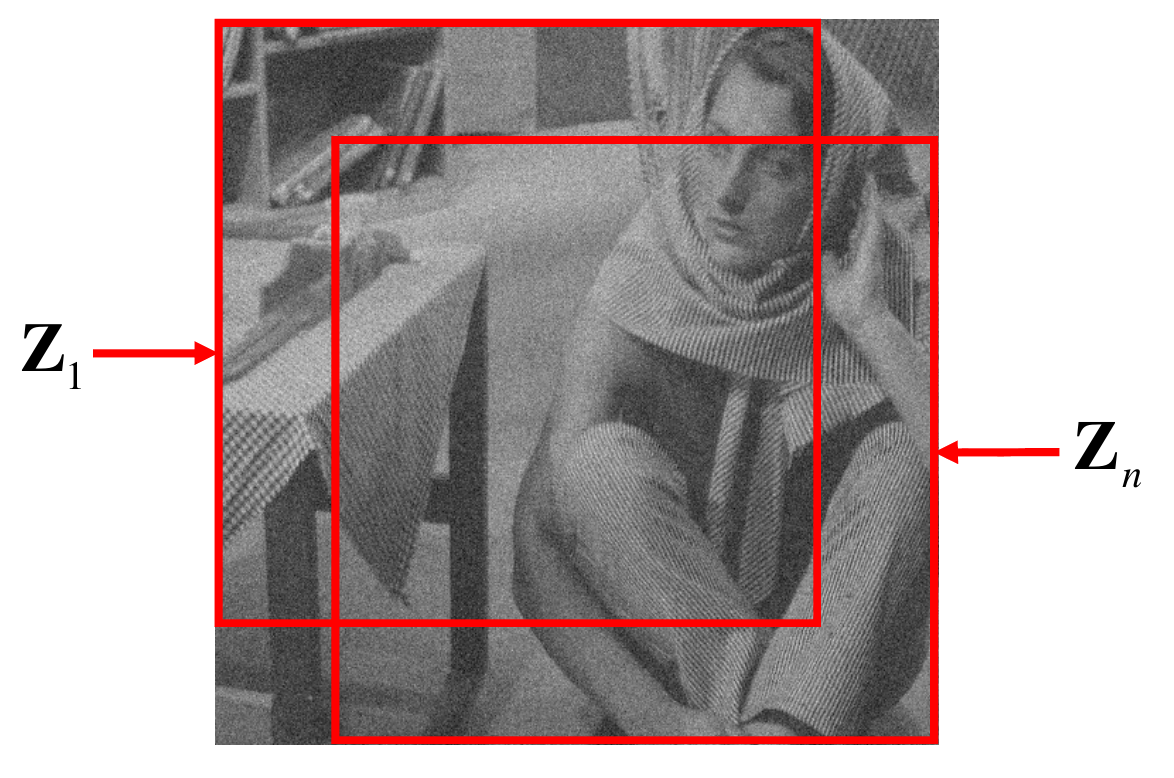} & \includegraphics[scale=0.2]{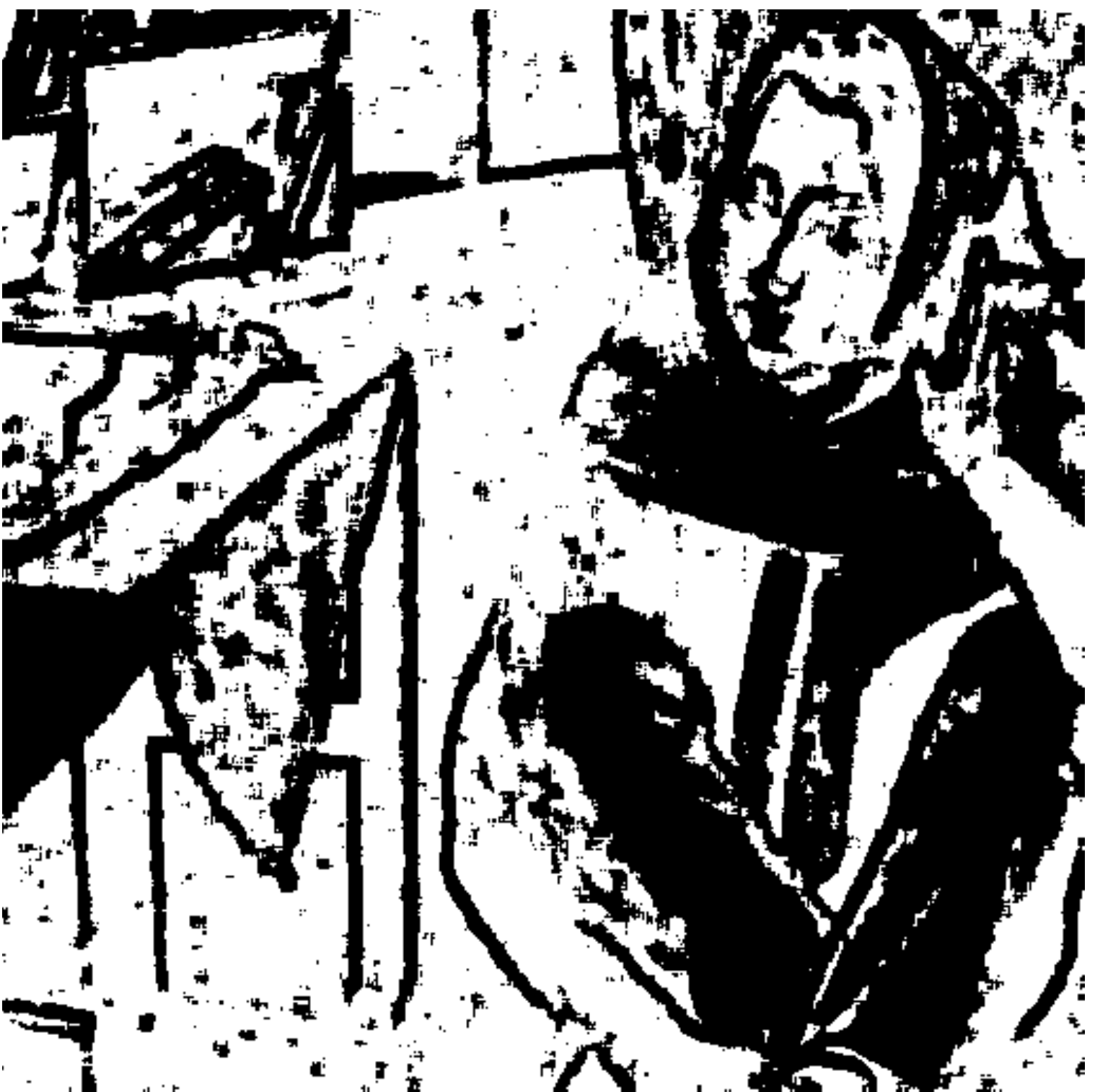}\\
{\small (a)} & {\small (b)}\\
\end{tabular}
\caption{(a) Two subimages $\mathbf{Z}_1$ and $\mathbf{Z}_n$ contained in a noisy version of the image Barbara.
(b) Classification of the pixels in a noisy version of the image Barbara to centers of smooth (white) and non-smooth (black) patches.}
\label{Figure: subimages}
\end{figure}

Let $\mathbf{X}$ be an $n\times(N_1-\sqrt{n}+1)(N_2-\sqrt{n}+1)$ matrix, containing
column stacked versions of all the $\sqrt{n}\times\sqrt{n}$ patches inside the image $\mathbf{Z}$.
We extract these patches column by column, starting from the top left-most patch.
When we calculated $\mathbf{P}$ as described in the previous section,
we assumed that each patch is associated only with its middle pixel.
Therefore $\mathbf{P}$ was designed to reorder the signal composed of the middle points in the patches,
which reside in the middle row of $\mathbf{X}$.
However, we can alternatively choose to associate all the patches with a pixel located in a different position,
e.g., the top left pixel in each patch.
This means that the matrix $\mathbf{P}$ can be used to reorder any one of the signals located in the rows of $\mathbf{X}$.
These signals are the column stacked versions of all the $n$ subimages of size $(N_1-\sqrt{n}+1)\times(N_2-\sqrt{n}+1)$ contained in the image $\mathbf{Z}$.
We denote these subimages by $\mathbf{Z}_j$, $j=1,2,\ldots,n$.
An example for two of them, $\mathbf{Z}_1$ and $\mathbf{Z}_n$,
contained in a noisy version of the image Barbara, is shown Fig \ref{Figure: subimages}(a).

We already observed in \cite{ram2011generalized} and \cite{ram2012redundant} that improved denoising results are obtained
when all the $n$ subimages of a noisy image are employed in its denoising process.
Here we use a similar scheme in order to improve the quality of the recovered image.
In order to avoid cumbersome notations we first describe a scheme which utilizes a single ordering matrix $\mathbf{P}$.
Let $\mathbf{z}_j=\mathbf{R}_j\mathbf{z}$ be the column stacked version of $\mathbf{Z}_j$,
where the matrix $\mathbf{R}_j$ extracts the $j$th subimage from the image $\mathbf{z}$.
We first calculate the matrix $\mathbf{P}$ using the patches in $\mathbf{X}$ and apply it to each subimage $\mathbf{z}_j$.
Then we apply the operator $H$ to each of the reordered subimages $\mathbf{z}_j^p=\mathbf{Pz}_j$,
apply the inverse permutation $\mathbf{P}^{-1}$ on the result, and obtain the reconstructed subimages
\begin{align}
\hat{\mathbf{y}}_j=\mathbf{P}^{-1}H\{\mathbf{P}\mathbf{z}_j\}=\mathbf{P}^{-1}H\{\mathbf{P}\mathbf{R}_j\mathbf{z}\}.
\label{rec y_j^p}
\end{align}
We next reconstruct the image from all the reconstructed subimages $\hat{\mathbf{y}}_j$ by plugging each subimage into its original place in the image canvas
and averaging the different values obtained for each pixel.
More formally, we obtain the reconstructed image $\hat{\mathbf{y}}$ as follows:
\begin{align}
\hat{\mathbf{y}}=\mathbf{D}^{-1}\sum_{j=1}^n\mathbf{R}_j^T\hat{\mathbf{y}}_j
=\mathbf{D}^{-1}\sum_{j=1}^n\mathbf{R}_j^T\mathbf{P}^{-1}H\{\mathbf{P}\mathbf{R}_j\mathbf{z}\}
\label{subimage rec1}
\end{align}
where the matrix $\mathbf{R}_j^T$ plugs the estimated $j$th subimage into its original place in the canvas,
and
\be
\mathbf{D}=\sum_{j=1}^n\mathbf{R}_j^T \mathbf{R}_j
\label{D}
\ee
is a diagonal weight matrix that simply averages the overlapping contributions per each pixel.
When $K$ random matrices $\mathbf{P}_k$ are employed,
we obtain the final estimate by averaging the images obtained with the different permutations
\begin{align}
\hat{\mathbf{y}}=\frac{1}{K}
\sum_{k=1}^{K}\left(\mathbf{D}^{-1}\sum_{j=1}^n\mathbf{R}_j^T\mathbf{P}_k^{-1}H\{\mathbf{P}_k\mathbf{R}_j\mathbf{z}\}\right).
\label{subimage rec}
\end{align}

This formula reveals two important properties of our scheme:
(i) the two summations that correspond to the two cycle-spinning versions lead to an averaging of $nK$ candidate solutions, a fact that boosts the overall performance of the recovery algorithm;
and (ii) if $H$ is chosen as linear, then the overall processing is linear as well, provided that we disregard the highly non-linear dependency of $\mathbf{P}$ on $\mathbf{z}$.

\subsection{Connection to BM3D}

The above processing scheme can be described a little differently.
We start by calculating the permutation matrix $\mathbf{P}$ from the image patches $\mathbf{x}_i$.
We then gather the patches by arranging them as the columns of a matrix $\mathbf{X}^p$ in the order defined by $\mathbf{P}$.
This matrix contains in its rows the reordered subimages $\mathbf{z}_j^p$,
therefore we next apply the operator $H$ to its rows,
and shuffle the columns of the resulting matrix according to the permutation defined by $\mathbf{P}^{-1}$.
We obtain a matrix $\tilde{\mathbf{X}}$, which contains in its rows the reconstructed subimages $\hat{\mathbf{y}}_j$,
and in its columns reconstructed versions $\tilde{\mathbf{x}}_i$ of the image patches $\mathbf{x}_i$.
We obtain the reconstructed image $\hat{\mathbf{y}}$ from the patches $\tilde{\mathbf{x}}_j$ by plugging them into their original places in the image canvas, and averaging the different values obtained for each pixel.
When $K$ random matrices $\mathbf{P}_k$ are employed, we apply the aforementioned scheme with each of these matrices,
and average the obtained images.

Now, looking at the image processing scheme described above, we can see some similarities to the first stage of the BM3D algorithm.
Both algorithms stack the image patches into groups, apply 1D processing across the patches,
return the patches into their place in the image, and average the results.
We note that the BM3D algorithm also applies a 2D transform to the patches before performing 1D processing across them.
This feature can be easily added to our scheme if needed, and we regard this as a preprocessing part of the operator $H$.
On the other hand, there are some key differences between the two schemes.
First, while the BM3D algorithm constructs a group  of neighbors for each patch,
here we order {\em all the patches} to one chain, which defines local neighbors.
Furthermore, this process is repeated $K$ times, implying that our approach consider $K$ different neighbors assignments.
Also, while in the BM3D the patch order in each group is not restricted,
ours is carefully determined as it plays a major role in our scheme.
Finally, the 1D processing applied by the BM3D consists of the use of a 1D transform, followed by thresholding and the inverse transform, implying a specific denoising.
Here we do not restrict ourselves to any specific 1D processing scheme, and allow the operator $H$ to be chosen based on the application at hand.
We next demonstrate our proposed schemes for image denoising and inpainting.

\section{Applications and Results}
\subsection{Image Denoising}

The problem of image denoising consists of the recovery of an image from its noisy version.
In that case $\mathbf{M}=\mathbf{I}$ and the corrupted image satisfies $\mathbf{z}=\mathbf{y}+\mathbf{v}$.
The patches $\mathbf{x}_i$ contain noise,
and we choose the distance measure between $\mathbf{x}_i$ and $\mathbf{x}_j$ to be the squared Euclidean distance divided by $n$, i.e
\begin{align}
w(\mathbf{x}_i,\mathbf{x}_j)=\frac{1}{n}\|\mathbf{x}_i-\mathbf{x}_j\|^2.
\label{Euclidean Distance}
\end{align}
In our previous works \cite{ram2011generalized} and \cite{ram2012redundant} we applied a complex multi-scale
processing on the ordered patches.
Here we wish to employ a far simpler scheme; we choose a 1D linear shift-invariant filter, and as we show next, we learn this filter from training images.  Furthermore, we suggest to switch between two such filters, based on the patch content.

We desire to treat smooth areas in the image differently than areas with edges or texture,
as our experiments show that this approach leads to better results.
More specifically, we employ different permutation matrices and filters in the smooth and non smooth areas of the image.
We first divide the patches into two sets:
$S_s$ - which contains smooth patches, and $S_e$ - which contains patches with edges or texture.
Let $\text{std}(\mathbf{x}_i)$ denote the standard deviation of the patch $\mathbf{x}_i$
and let $C$ be a scalar design parameter.
Then we use the following classification rule: if $\text{std}(\mathbf{x}_i)<C\sigma$ then $\mathbf{x}_i\in S_s$,
otherwise $\mathbf{x}_i\in S_e$.
Fig \ref{Figure: subimages}(b) demonstrates the application of this classification rule to the noisy Barbara image shown in Fig. \ref{Figure: subimages}(a),
where we use the parameters $\sqrt{n}=8$ and $C=1.2$ which we will later use in the denoising process of this image.
White pixels are the centers of smooth patches and black pixels are the centers of patches containing texture or edges.
It can be seen that the obtained image indeed contains a rough classification of the patches into smooth and non smooth sets.

We next divide each subimage $\mathbf{z}_j$ into two signals: $\mathbf{z}_{j,s}$ - which contains the pixels corresponding to the  smooth patches, and $\mathbf{z}_{j,e}$ - which contains the pixels corresponding to the patches with edges and texture.
We apply the nearest neighbors search method described above to the patches in the sets $S_s$ and $S_e$,
and obtain two different permutation matrices $\mathbf{P}_s$ and $\mathbf{P}_e$.
$\mathbf{P}_s$ and $\mathbf{P}_e$ extract from $\mathbf{z}_j$ the signals $\mathbf{z}_{j,s}$ and $\mathbf{z}_{j,e}$, respectively, and then each apply a different permutation.
We apply $\mathbf{P}_s$ and $\mathbf{P}_e$ to $\mathbf{z}_{j,s}$ and $\mathbf{z}_{j,e}$
and obtain $\mathbf{z}_{j,s}^p$ and $\mathbf{z}_{j,e}^p$, respectively, which are the signals to which we apply the filters.
More formally,
\begin{align}
\begin{bmatrix}
\mathbf{z}_{j,s}^p\\
\mathbf{z}_{j,e}^p
\end{bmatrix}
=
\begin{bmatrix}
P_s\{\mathbf{z}_{j,s}\}\\
P_e\{\mathbf{z}_{j,e}\}\\
\end{bmatrix}
=
\begin{bmatrix}
\mathbf{P}_s\\
\mathbf{P}_e
\end{bmatrix}
\mathbf{z}_j
=\mathbf{Pz}_j=\mathbf{z}_j^p
\end{align}
where we defined the matrix
\be
\mathbf{P}=\begin{bmatrix}
\mathbf{P}_s\\
\mathbf{P}_e
\end{bmatrix}.
\ee

We next wish to find the filters $\mathbf{h}_s$ and $\mathbf{h}_e$
applied to $\mathbf{z}_{j,s}^p$ and $\mathbf{z}_{j,e}^p$, respectively.
We denote the convolution matrices corresponding to $\mathbf{z}_{j,s}^p$ and $\mathbf{z}_{j,e}^p$ by $\mathbf{Z}_{j,s}^p$ and $\mathbf{Z}_{j,e}^p$,
and obtain the filtered subimages
\begin{align}
\hat{\mathbf{y}}_{j}&=
\mathbf{P}^{-1}
\begin{bmatrix}
\mathbf{Z}_{j,s}^p\mathbf{h}_s\\
\mathbf{Z}_{j,e}^p\mathbf{h}_e
\end{bmatrix}
=\mathbf{P}^{-1}\begin{bmatrix}
\mathbf{Z}_{j,s}^p & 0\\
0 & \mathbf{Z}_{j,e}^p
\end{bmatrix}
\begin{bmatrix}
\mathbf{h}_s\\
\mathbf{h}_e
\end{bmatrix}\nonumber\\
&=\mathbf{P}^{-1}\mathbf{Z}_j^p\mathbf{h}
\label{filtered_y_j}
\end{align}
where we defined
\be
\mathbf{Z}_j^p=\begin{bmatrix}
\mathbf{Z}_{j,s}^p & 0\\
0 & \mathbf{Z}_{j,e}^p
\end{bmatrix},
\mathbf{h}=
\begin{bmatrix}
\mathbf{h}_s\\
\mathbf{h}_e
\end{bmatrix}.
\ee
The vector $\mathbf{h}$ stores the filter taps to be designed.
We substitute (\ref{filtered_y_j}) in (\ref{subimage rec1}), and obtain the reconstructed image
\begin{align}
\hat{\mathbf{y}}&=\mathbf{D}^{-1}\sum_{j=1}^n\mathbf{R}_j^T\mathbf{P}^{-1}\mathbf{Z}_j^p\mathbf{h}.
\end{align}
When $K$ random matrices $\mathbf{P}_k$ are employed, we obtain the final estimate by averaging the images obtained with the different matrices
\begin{align}
\hat{\mathbf{y}}&=\frac{1}{K}\sum_{k=1}^{K}\left(\mathbf{D}^{-1}\sum_{j=1}^n\mathbf{R}_j^T\mathbf{P}_k^{-1}\mathbf{Z}_j^p\mathbf{h}\right)\nonumber\\
&=\frac{1}{K}\sum_{k=1}^{K}\mathbf{D}^{-1}[\mathbf{R}_1^T,\ldots,\mathbf{R}_n^T]
\begin{bmatrix}
\mathbf{P}_k^{-1}\mathbf{Z}_1^p\\
\vdots\\
\mathbf{P}_k^{-1}\mathbf{Z}_n^p
\end{bmatrix}
\mathbf{h}
=\mathbf{Qh}
\end{align}
where we defined \be
\mathbf{Q}=\frac{1}{K}\sum_{k=1}^{K}\mathbf{D}^{-1}[\mathbf{R}_1^T,\ldots,\mathbf{R}_n^T]
\begin{bmatrix}
\mathbf{P}_k^{-1}\mathbf{Z}_1^p\\
\vdots\\
\mathbf{P}_k^{-1}\mathbf{Z}_n^p
\end{bmatrix}.
\label{Q}
\ee

Now let $\mathbf{y}^g$, $j=1,\ldots,G$ be a training set which contains the column stack versions of $G$ clean images.
For each such image we create a noisy version $\mathbf{z}^g$ by adding it noise with the same statistics as the noise in $\mathbf{z}$.
Then we calculate for each image $\mathbf{z}^g$ a matrix $\mathbf{Q}^g$ using (\ref{Q}),
and learn the filters vector $\mathbf{h}$ by minimizing
\begin{align}
\hat{\mathbf{h}}&=\underset{\mathbf{h}}{\operatorname{argmin}}
\sum_{g=1}^G\|\mathbf{y}^g-\mathbf{Q}^g\mathbf{h}\|^2\nonumber\\
&=\left[\sum_{g=1}^G (\mathbf{Q}^g)^T \mathbf{Q}^g\right]^{-1}\sum_{k=1}^G (\mathbf{Q}^g)^T \mathbf{y}^g.
\end{align}
Once we have the filters vector $\hat{\mathbf{h}}$ we can employ it to denoise $\mathbf{z}$ by building $\mathbf{Q}$ using (\ref{Q})
and then calculating
\be
\hat{\mathbf{y}}=\mathbf{Q}\hat{\mathbf{h}}.
\ee
We can further improve our results by applying a second iteration of our proposed scheme,
in which all the processing stages remain the same, but the permutation matrices are built using patches extracted from the first iteration clean result.

\begin{figure}[t]
\centering
\begin{tabular}{cc}
\includegraphics[scale=0.22]{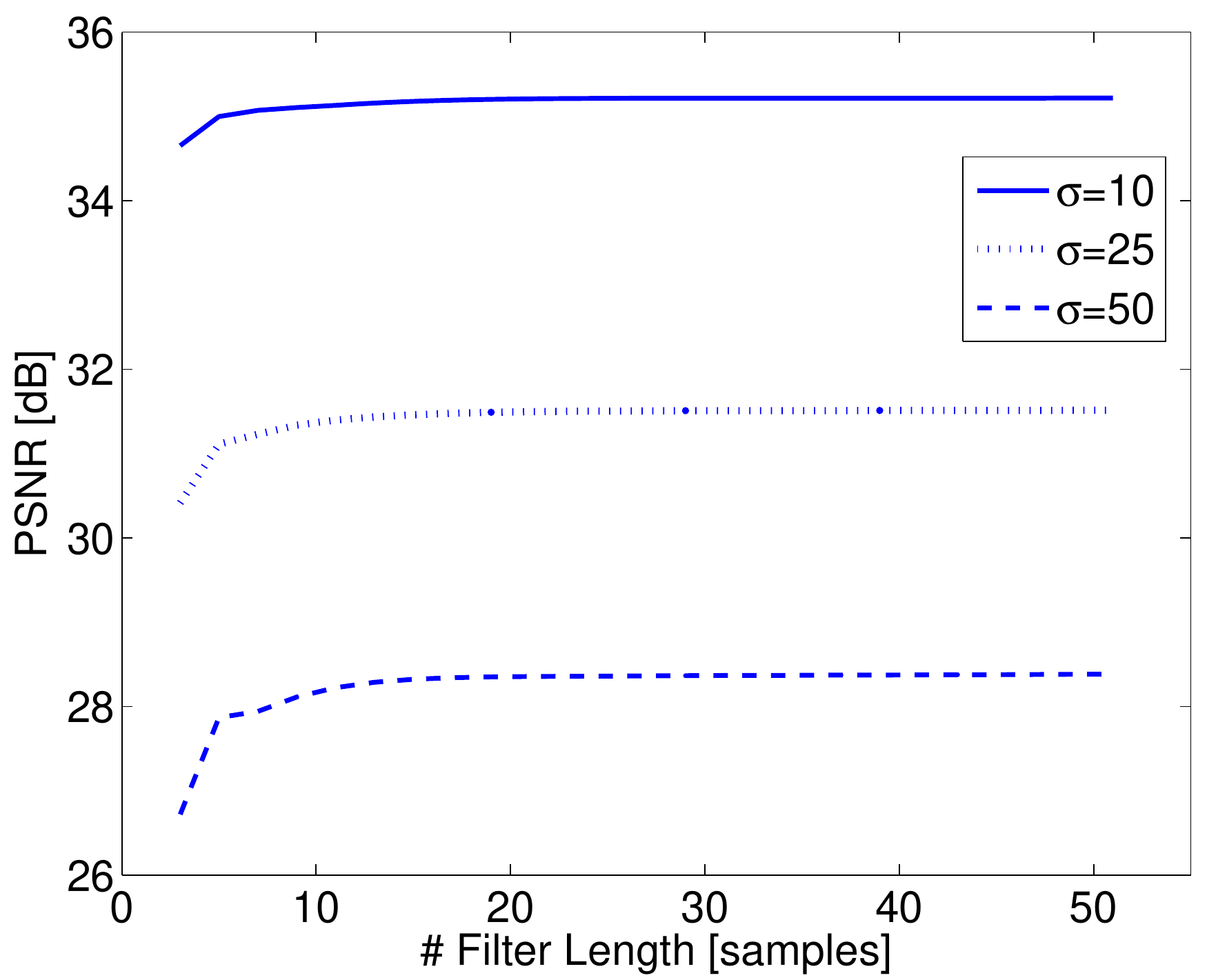} & \includegraphics[scale=0.22]{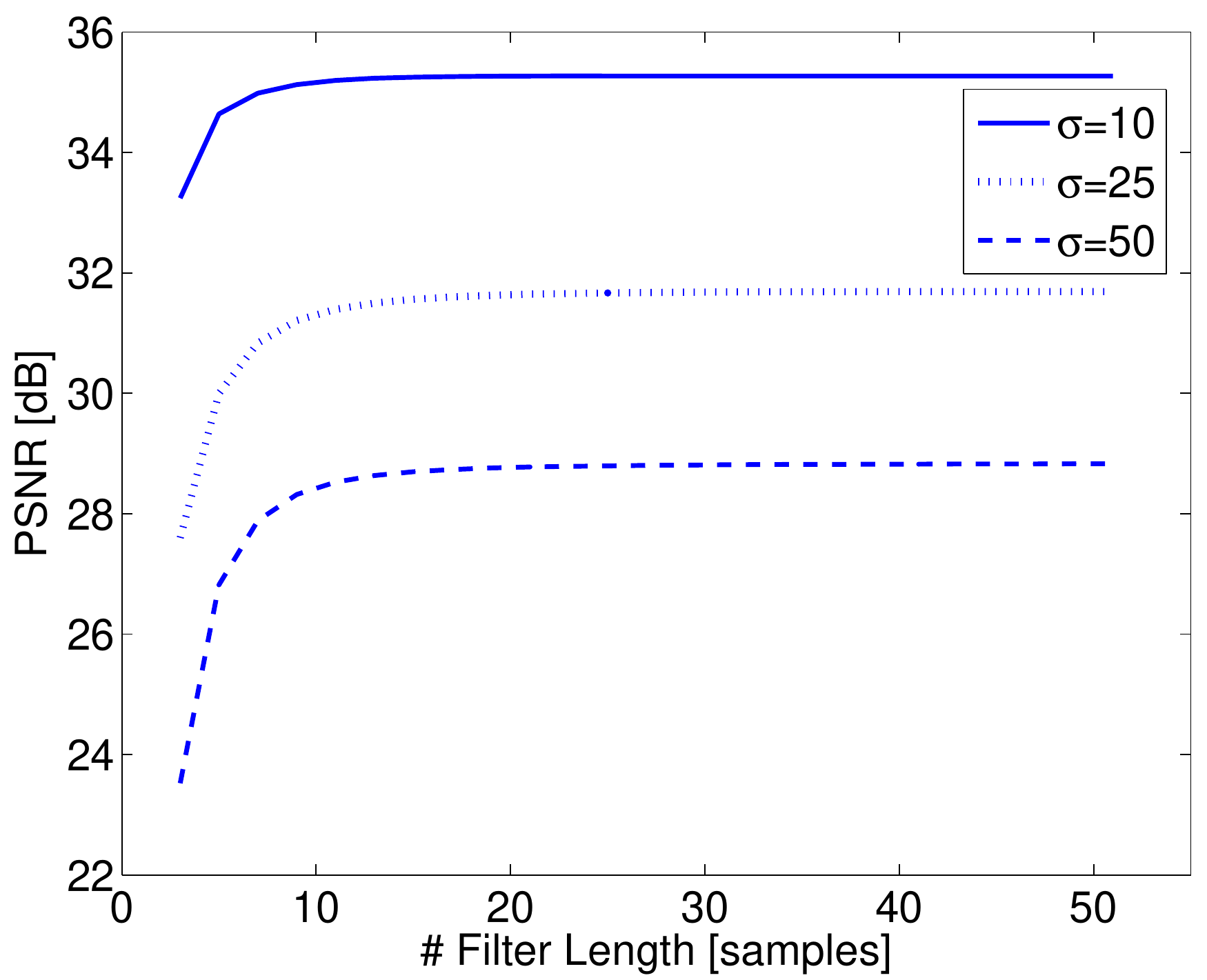}\\
\end{tabular}
\caption{Average of the PSNR values obtained for the 3 test images, as a function of the filter length, for the different noise levels: (a) First iteration. (b) Second Iteration.}
\label{Figure: filters length}
\end{figure}

\begin{figure}[t]
\centering
\begin{tabular}{cc}
\includegraphics[scale=0.22]{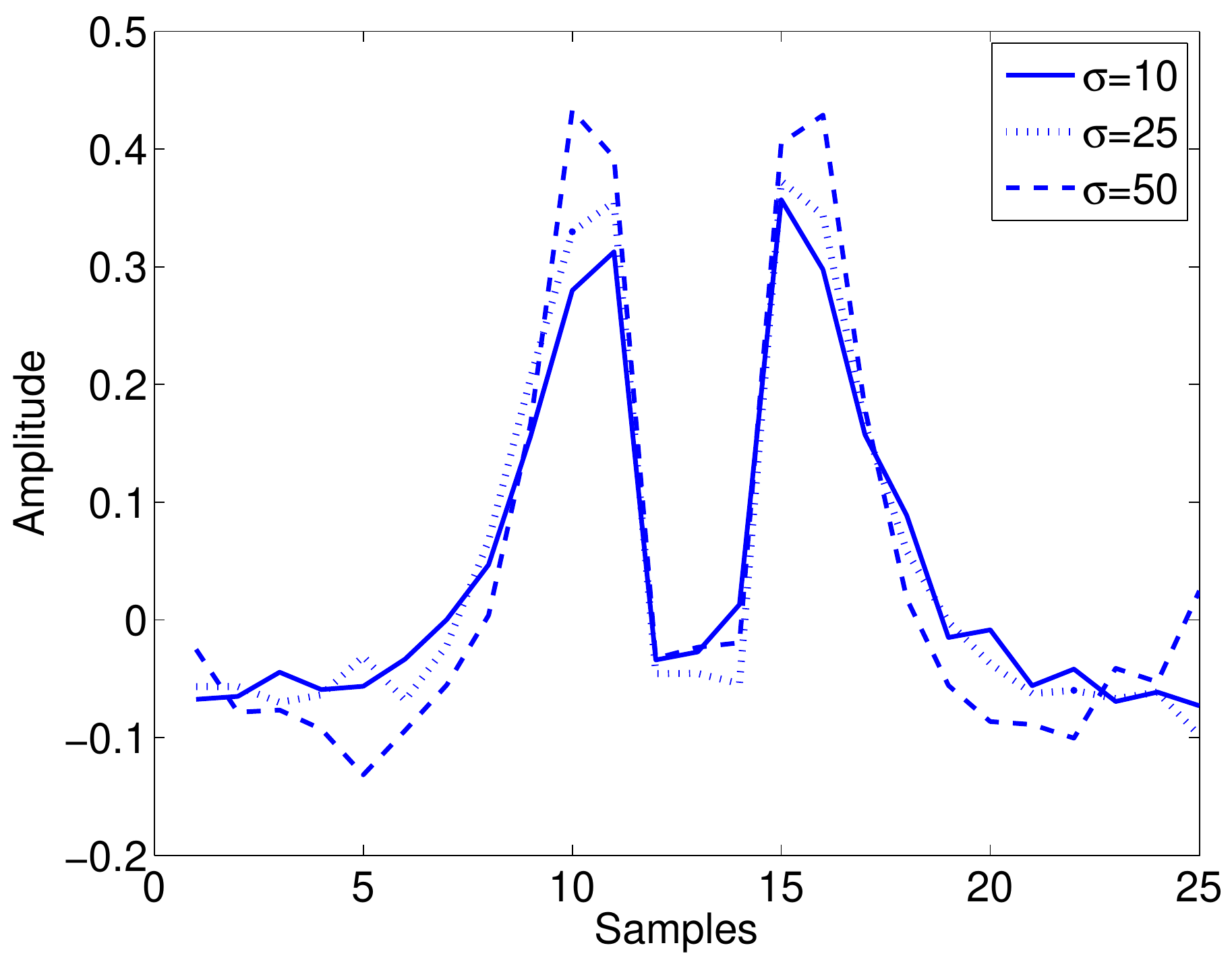} & \includegraphics[scale=0.22]{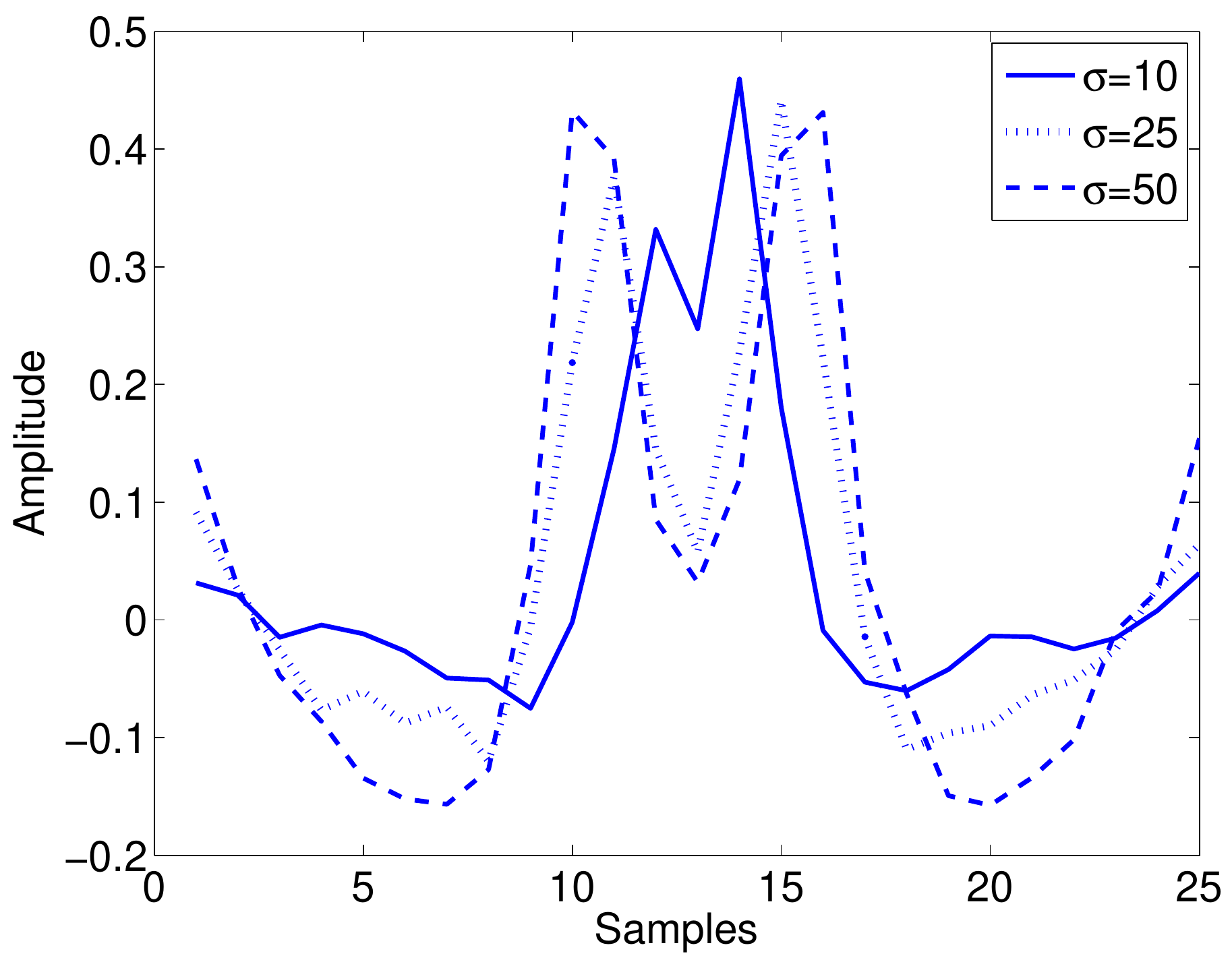}\\
\includegraphics[scale=0.22]{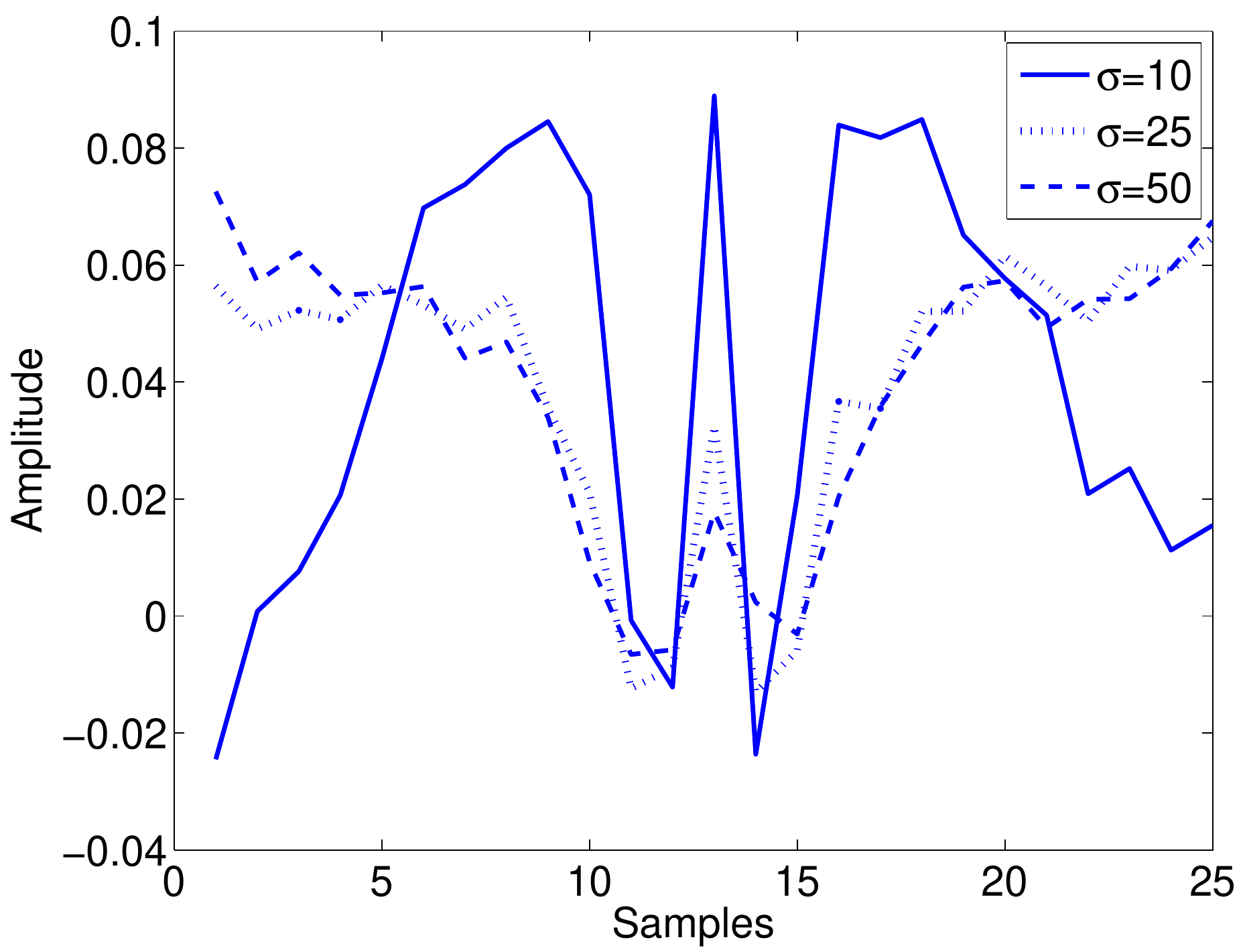} & \includegraphics[scale=0.22]{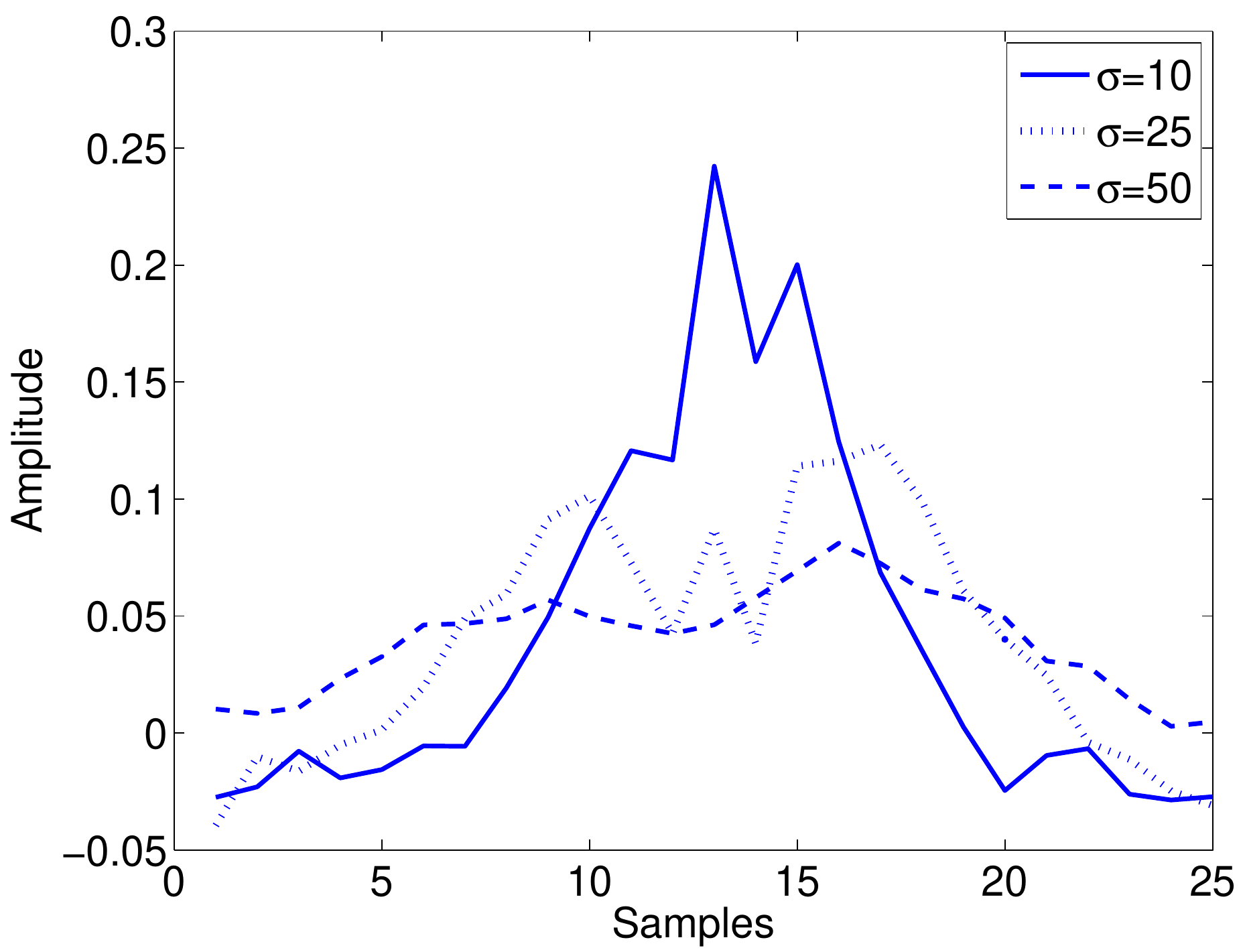}\\
\end{tabular}
\caption{The trained filters learned from noisy images:
Left column - the filters $\mathbf{h}_s$ obtained in the first (top) and second (bottom) iterations.
Right column - the filters $\mathbf{h}_e$ obtained in the first (top) and second (bottom) iterations.}
\label{Figure: trained filters}
\end{figure}

In order to assess the performance of the proposed image denoising scheme
we apply it to a test set containing noisy versions of the images Lena, Barbara and House,
with noise standard deviations $\sigma=10,25,50$.
We learn the filters vector $\mathbf{h}$ from a training set containing the images Man, Peppers, Boat and Fingerprint.
The parameters employed by the proposed denoising scheme for the three noise levels are shown in Table \ref{Table: denoising parameters}.
We note that the reason we chose a uniform filter length of 25 samples for all noise levels can be justified using Fig. \ref{Figure: filters length}.
Fig. \ref{Figure: filters length}. shows the average of the PSNR values obtained in the first and second iterations for the 3 test images, as a function of the filter length, for the different noise levels.
It can be seen that in both iterations the performance gain obtained using filters longer than 25 samples is negligible.
The trained filters obtained in each iteration for the different noise levels are shown in Fig. \ref{Figure: trained filters}.
First, it can be seen that filters $\mathbf{h}_s$ and $\mathbf{h}_e$ indeed look different.
It can also be seen that in the first iteration the shape of the filter $\mathbf{h}_s$ does not change much as the noise level increases, and in the second iteration the filters obtained for the higher noise levels are similar, but very different from the filter obtained for $\sigma=10$.
On the other hand, in both iterations the shape of the filter $\mathbf{h}_e$ changes greatly as the noise level increases.

\begin{figure*}[t]
\centering
\begin{tabular}{ccc}
\includegraphics[scale=0.215]{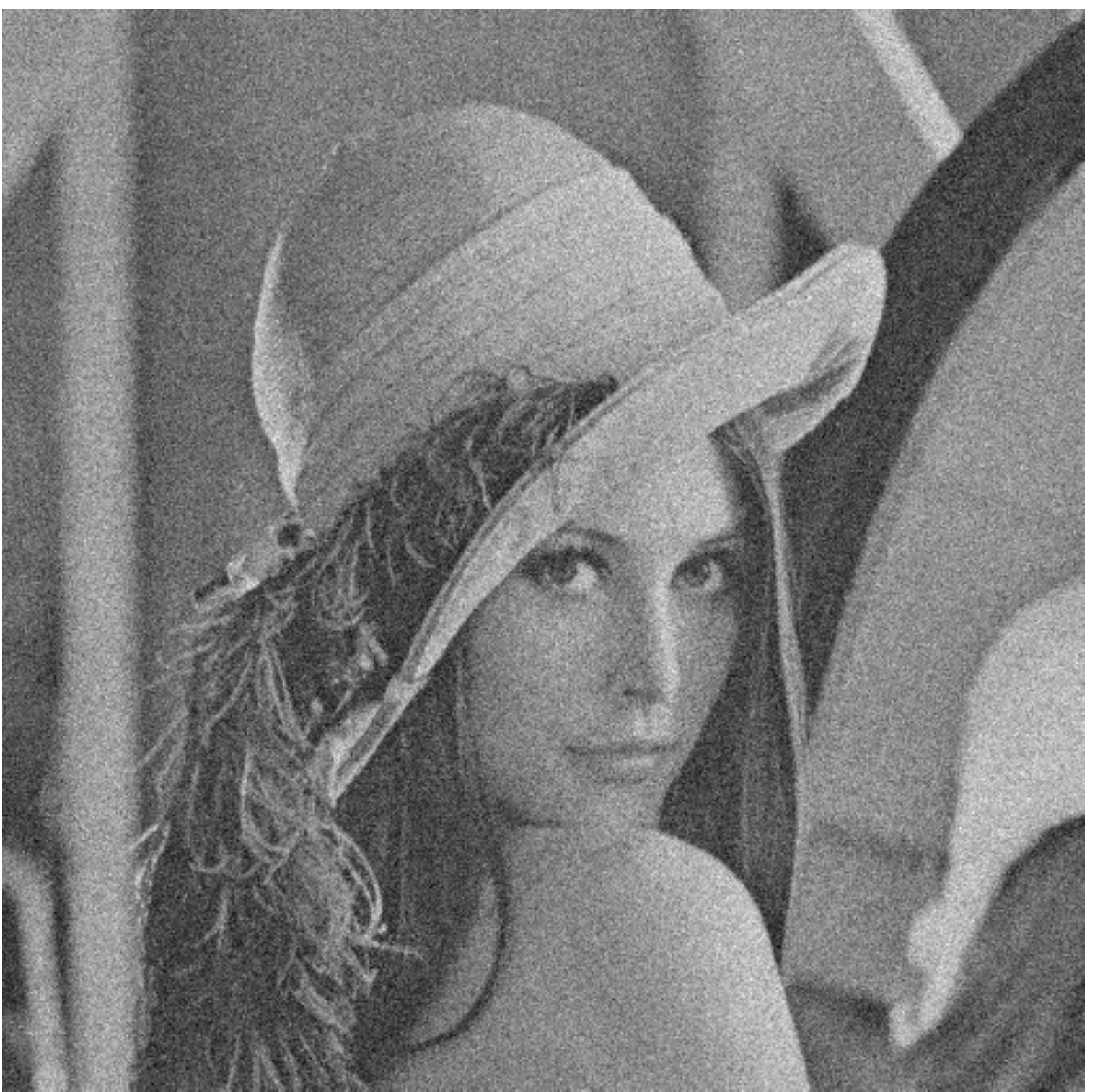} & \includegraphics[scale=0.215]{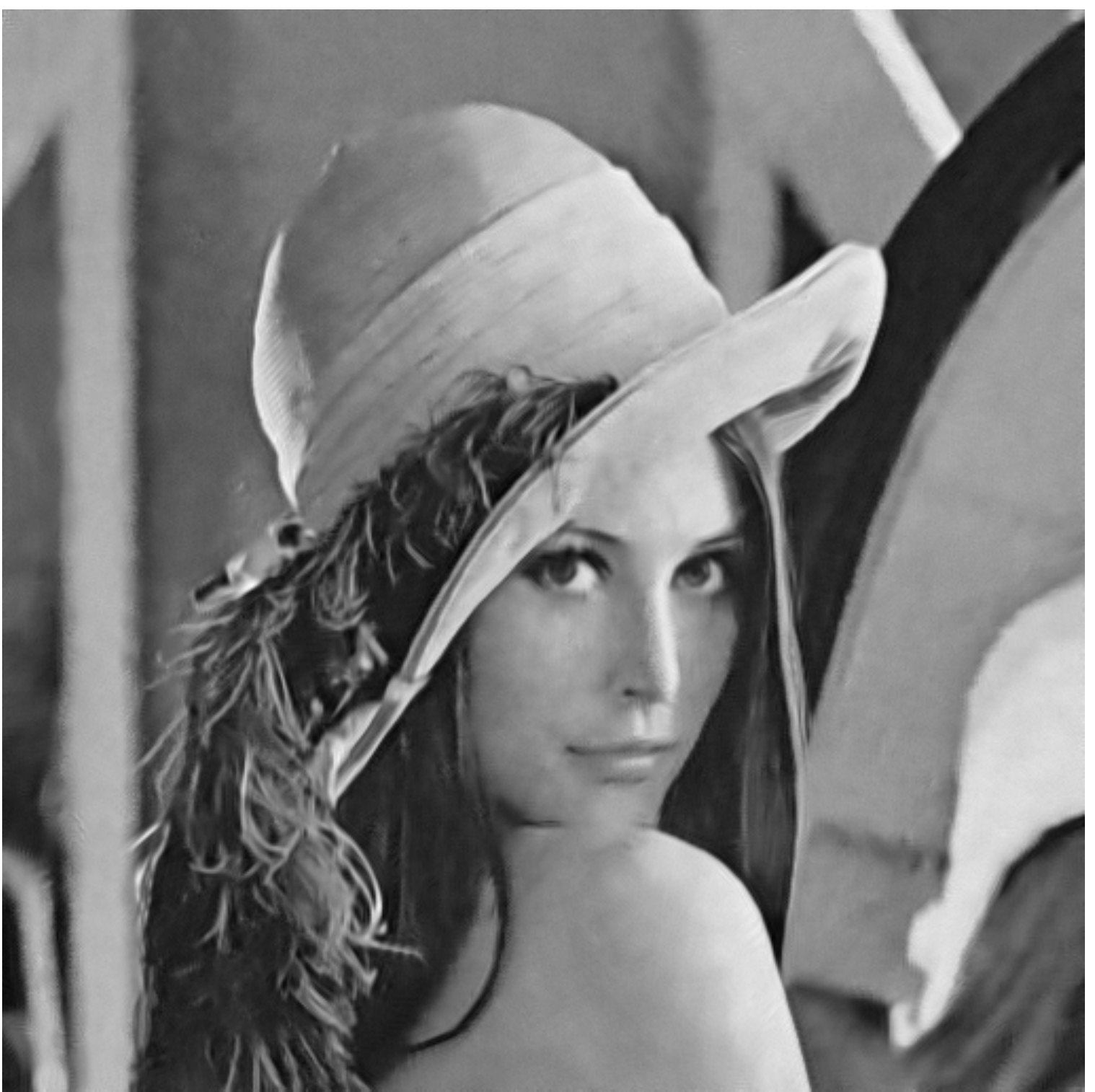}
& \includegraphics[scale=0.215]{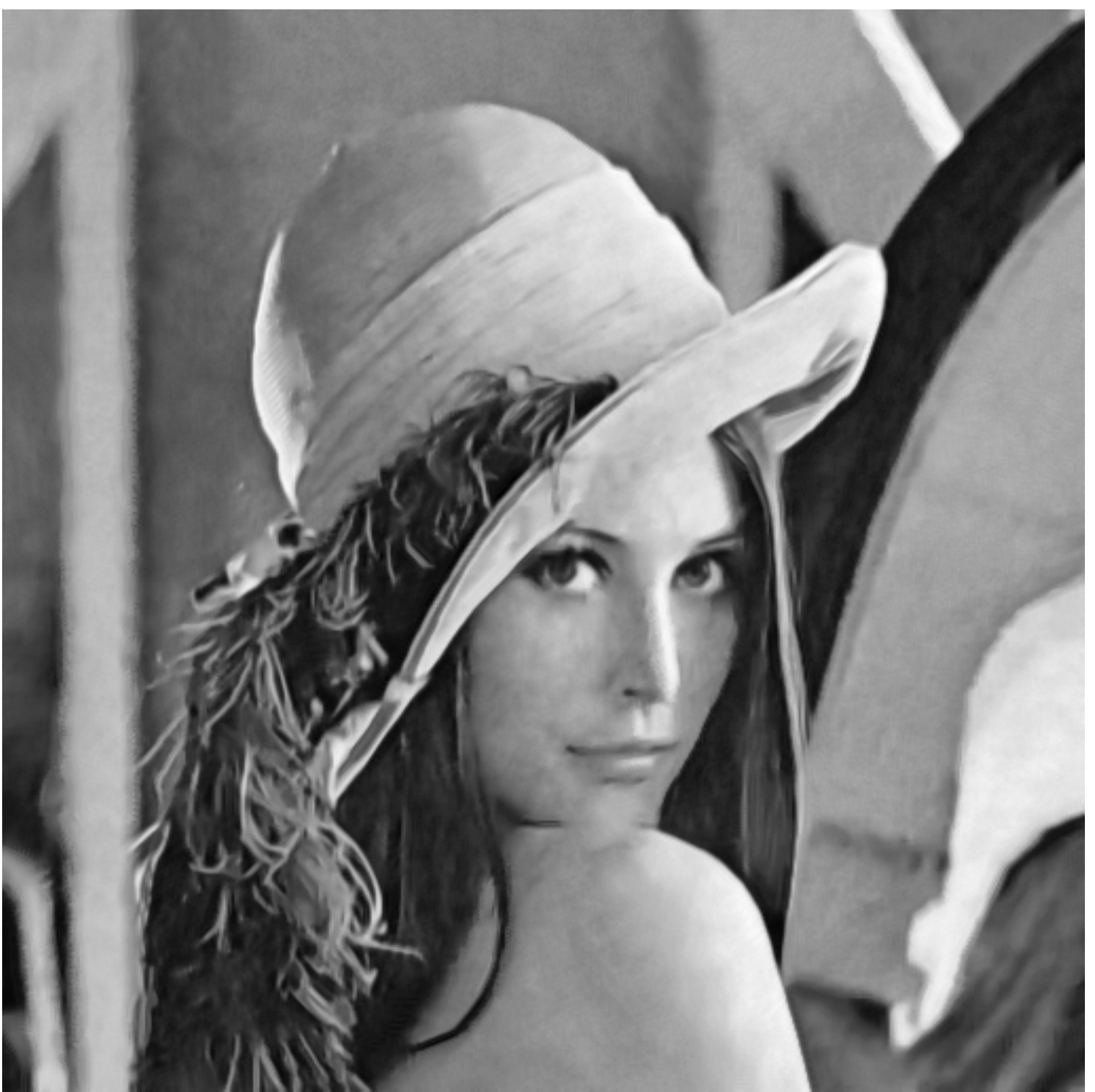}\\
& {PSNR=31.58 dB} & {PSNR=31.81 dB}\\
\includegraphics[scale=0.215]{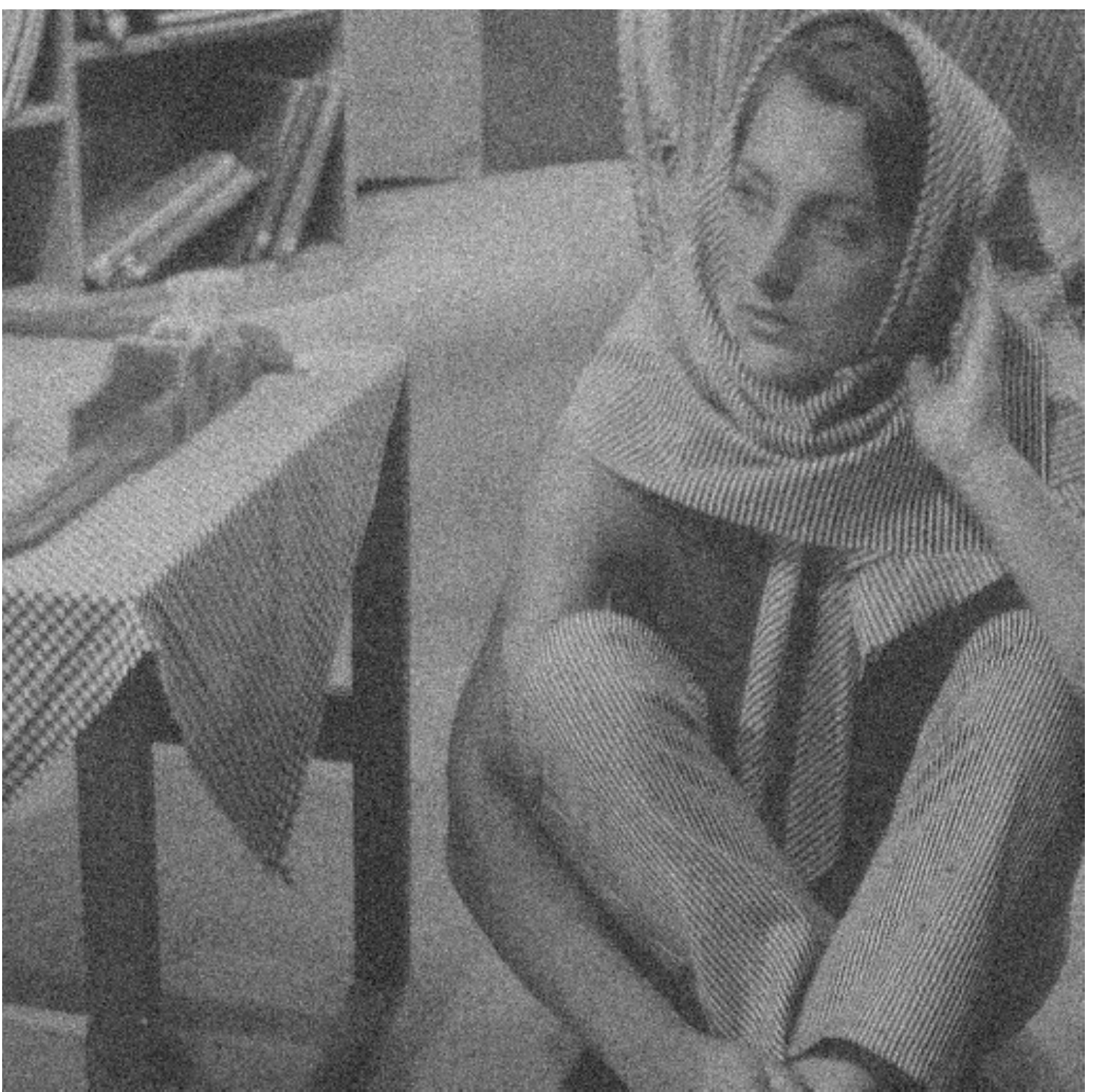} & \includegraphics[scale=0.215]{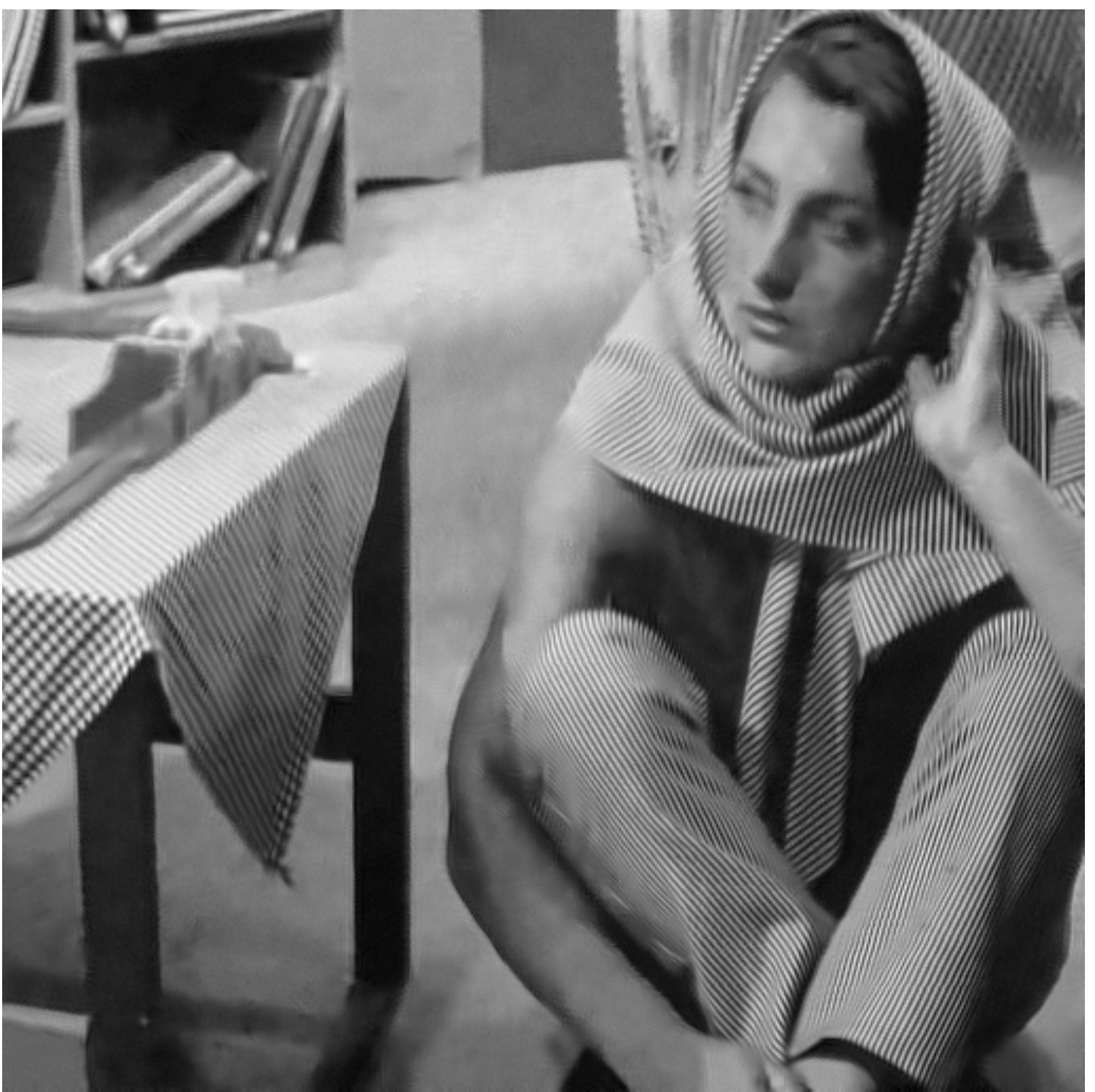}
& \includegraphics[scale=0.215]{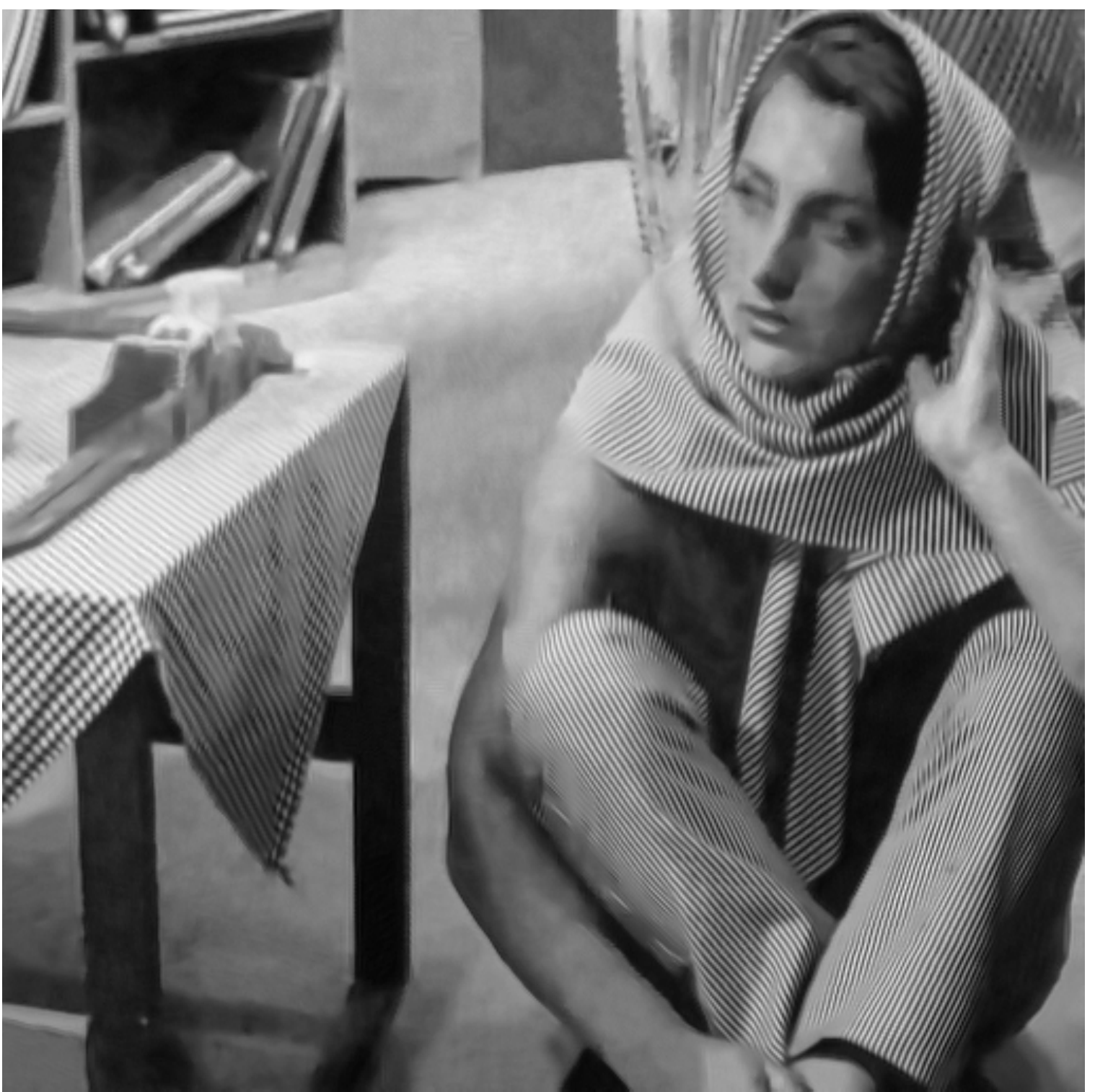}\\
& {PSNR=30.46 dB} & {PSNR=30.54 dB}\\
\includegraphics[scale=0.215]{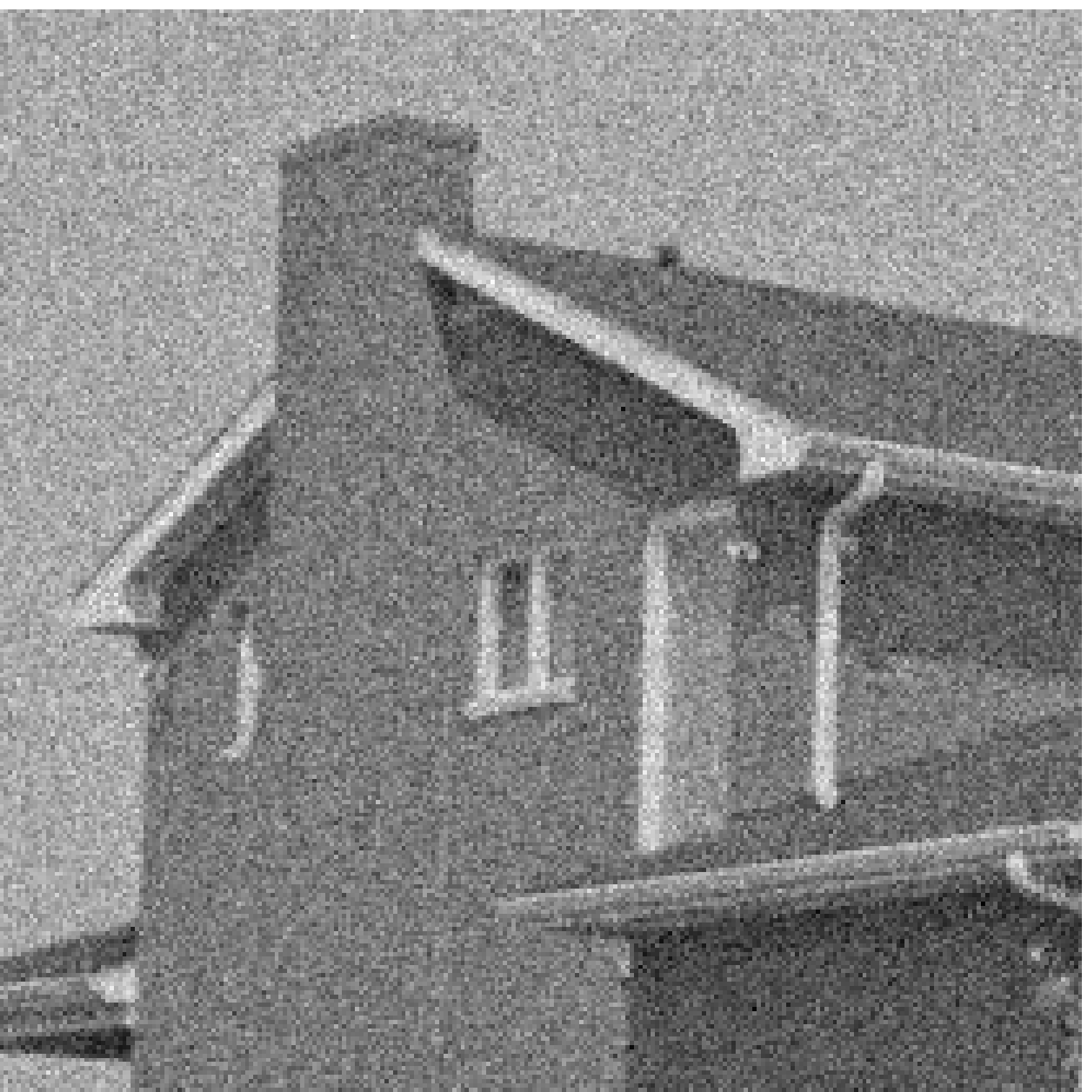} & \includegraphics[scale=0.215]{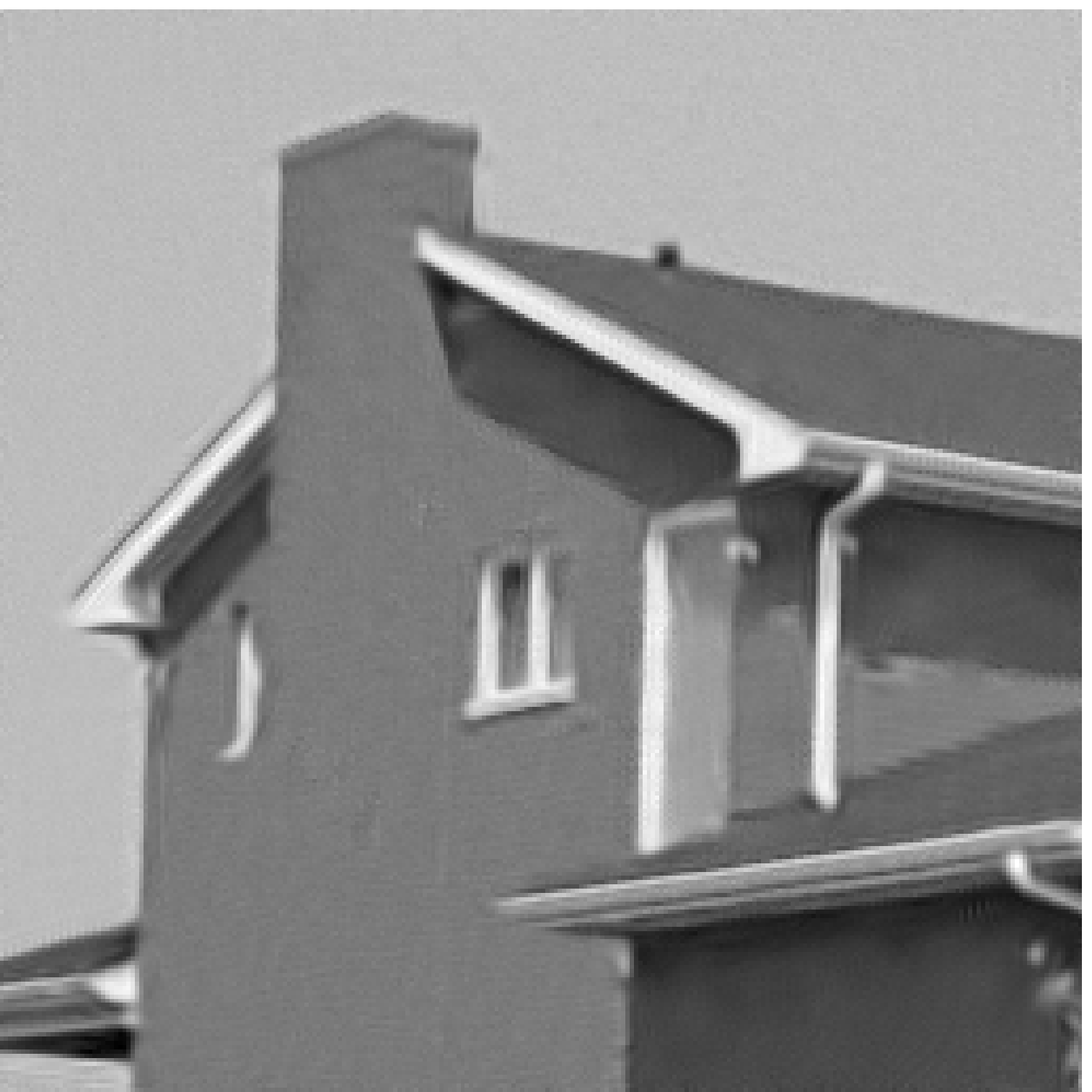}
& \includegraphics[scale=0.215]{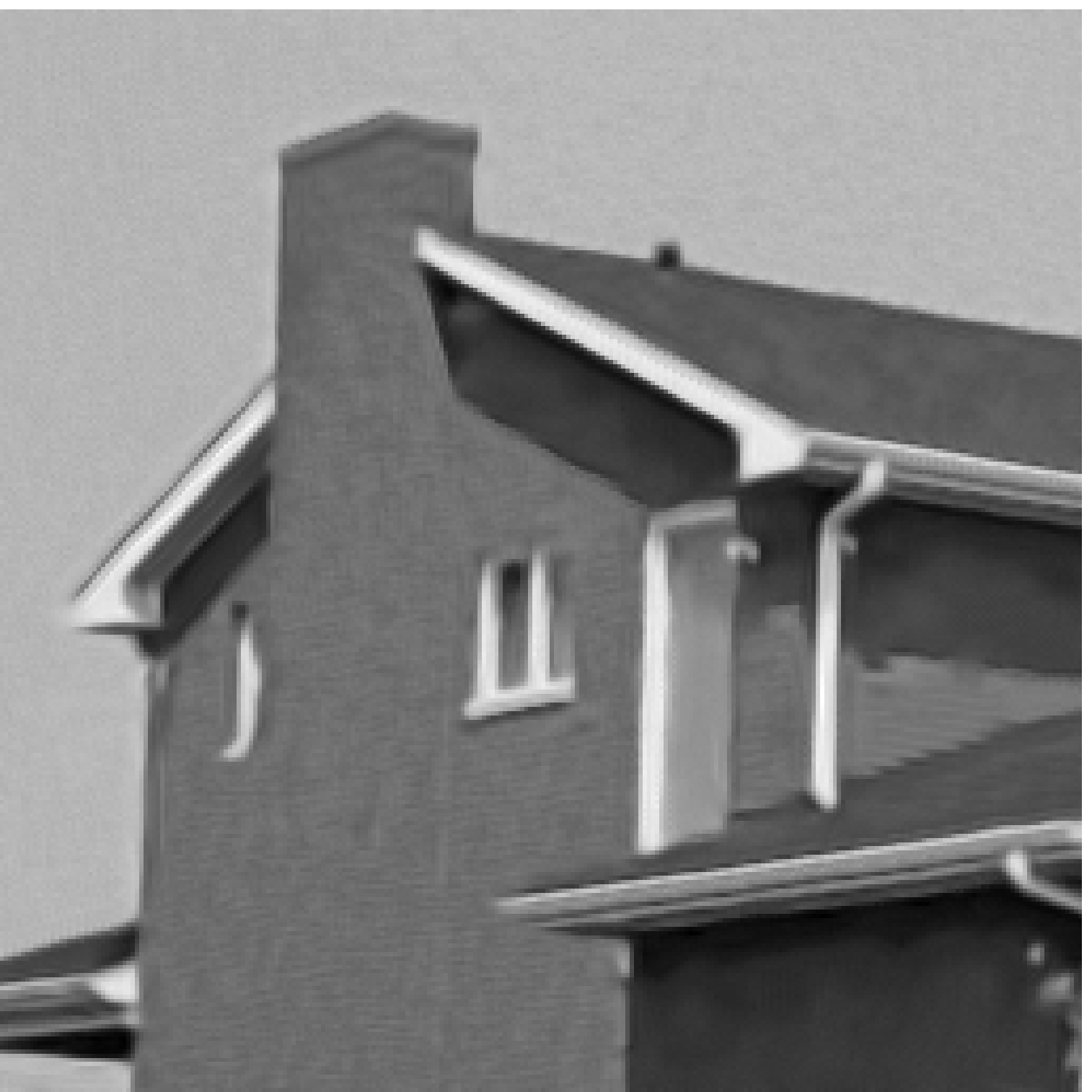}\\
& {PSNR=32.48 dB} & {PSNR=32.65 dB}\\
\end{tabular}
\caption{Denoising results (PSNR) for the images Lena, Barbara and House ($\sigma = 25$, input PSNR=20.18 dB):
Left column - noisy images, Center column - 1 iteration results, and Right column - 2 iterations results.}
\label{Figure: denoised images}
\end{figure*}

For comparison, we also apply the K-SVD algorithm \cite{elad2006image}.
The PSNR values of the results obtained with this algorithm, and two iterations of our denoising scheme are shown in Table \ref{Table: full_images_PSNR}.
The noisy and recovered images obtained with our scheme for $\sigma=25$ are shown in Fig. \ref{Figure: denoised images}.
First, it can be seen that the second iteration improves the results of our proposed scheme in all the cases but one.
It can also be seen that the results obtained with two iterations of our scheme are comparable to the ones of the K-SVD for $\sigma=10$, but are much better than the ones of the K-SVD for $\sigma=25$ and $\sigma=50$.

\begin{table}[t]
\centering
\caption{Parameters used in the denoising experiments.\newline}
\begin{tabular}{|c|c|c|c|c|c|c|c|}\hline
$\sigma$ & Iteration & $K$ & $\sqrt{n}$ & $B$ & $C$ & $\epsilon$ & Filters Length\\\hline
\multirow{2}{*}{10} & 1 & 10 & 6 & 111 & 1.6 & $10^6$ & 25\\\cline{2-8}
& 2 & 10 & 4 & 441 & 0.8 & $10^6$ & 25\\\hline
\multirow{2}{*}{25} & 1 & 10 & 8 & 111 & 1.2 & $10^6$ & 25\\\cline{2-8}
& 2 & 10 & 4 & 441 & 0.4 & $10^6$ & 25\\\hline
\multirow{2}{*}{50} & 1 & 10 & 12 & 111 & 1.1 & $10^6$ & 25\\\cline{2-8}
& 2 & 10 & 5 & 441 & 0.2 & $10^6$ & 25\\\hline
\end{tabular}
\label{Table: denoising parameters}
\end{table}

\begin{table}[t]
\centering
\caption{Denoising results (PSNR in dB) of noisy versions of
the images Lena, Barbara and House,obtained with the K-SVD algorithm and the proposed scheme.
For each image and noise level the best result is highlighted.\newline}
\begin{tabular}{|c|c|c|c|c|}\hline
\multirow{2}{*}{Image} & \multirow{2}{*}{Method} & \multicolumn{3}{c|}{$\sigma$/PSNR}\\\cline{3-5}
& & $10/28.14$ & $25/20.18$ & $50/14.16$ \\\hline
\multirow{3}{*}{Lena} & K-SVD & $\mathbf{35.49}$  & 31.36 & 27.82 \\\cline{2-5}
& proposed (1 iter.) & 35.33 & 31.58 & 28.54 \\\cline{2-5}
& proposed (2 iter.) & 35.41 & $\mathbf{31.81}$ & $\mathbf{29.00}$ \\\noalign{\hrule height 1pt}
\multirow{3}{*}{Barbara} & K-SVD & 34.41  & 29.53 &  25.4\\\cline{2-5}
& proposed (1 iter.)  & $\mathbf{34.48}$ &  30.46 &  27.17  \\\cline{2-5}
& proposed (2 iter.) & 34.46 & $\mathbf{30.54}$ & $\mathbf{27.45}$ \\\noalign{\hrule height 1pt}
\multirow{3}{*}{House} & K-SVD & $\mathbf{36.00}$  & 32.12 &  28.15 \\\cline{2-5}
& proposed (1 iter.)  & 35.83 & 32.48 & 29.37 \\\cline{2-5}
& proposed (2 iter.) & 35.94 & $\mathbf{32.65}$ & $\mathbf{29.93}$ \\\hline
\end{tabular}
\label{Table: full_images_PSNR}
\end{table}

\subsection{Image Inpainting}

The problem of image inpainting consists of the recovery of missing pixels in the given image.
Here we handle the case where there is no additive noise,
therefore $\mathbf{v}=0$, and $\mathbf{M}$ is a diagonal matrix of size $N\times N$ which contains ones and zeroes in its main diagonal corresponding to existing and missing pixels, correspondingly.
Each patch may contain missing pixels, and we denote by $S_i$ the set of indices of non-missing pixels in the patch $\mathbf{x}_i$.
We choose the distance measure between patches $\mathbf{x}_i$ and $\mathbf{x}_j$ to be
the average of squared differences between existing pixels that share the same location in both patches, i.e.
\begin{equation}
w(\mathbf{x}_i,\mathbf{x}_j)=\frac{\sum_{k\in S_i\cap S_j}(x_i[k]-x_j[k])^2}{|S_i\cap S_j|}.
\label{Inpainting Distance}
\end{equation}

We start by calculating the matrix $\mathbf{P}$ according to the scheme described in Section \ref{smooth P},
with a minor difference: when a patch does not share pixels with any of the unvisited patches,
the next patch in the path is chosen to be its nearest spatial neighbor.
We next apply the obtained matrix to the subimages $\mathbf{z}^j$,
and observe that the permuted vectors $\mathbf{z}_j^p=\mathbf{P}\mathbf{z}^j$ contain missing values.
We bear in mind that the target signals $\mathbf{y}_j^p=\mathbf{P}\mathbf{y}^j$ should be smooth,
and therefore apply on the subimages $\mathbf{z}_j^p$ an operator $H$ which recovers the missing values using cubic interpolation.
We apply the matrix $\mathbf{P}^{-1}$ on the resulting vectors and obtain the estimated subimages $\hat{\mathbf{y}}_j$.
The final estimate is obtained from these subimages using (\ref{subimage rec1}).
We improve our results by applying two additional iterations of a modified version of this inpainting scheme,
where the only difference is that we rebuild $\mathbf{P}$ using reconstructed (and thus full patches).

\begin{figure*}[t]
\centering
\begin{tabular}{ccccc}
\includegraphics[scale=0.205]{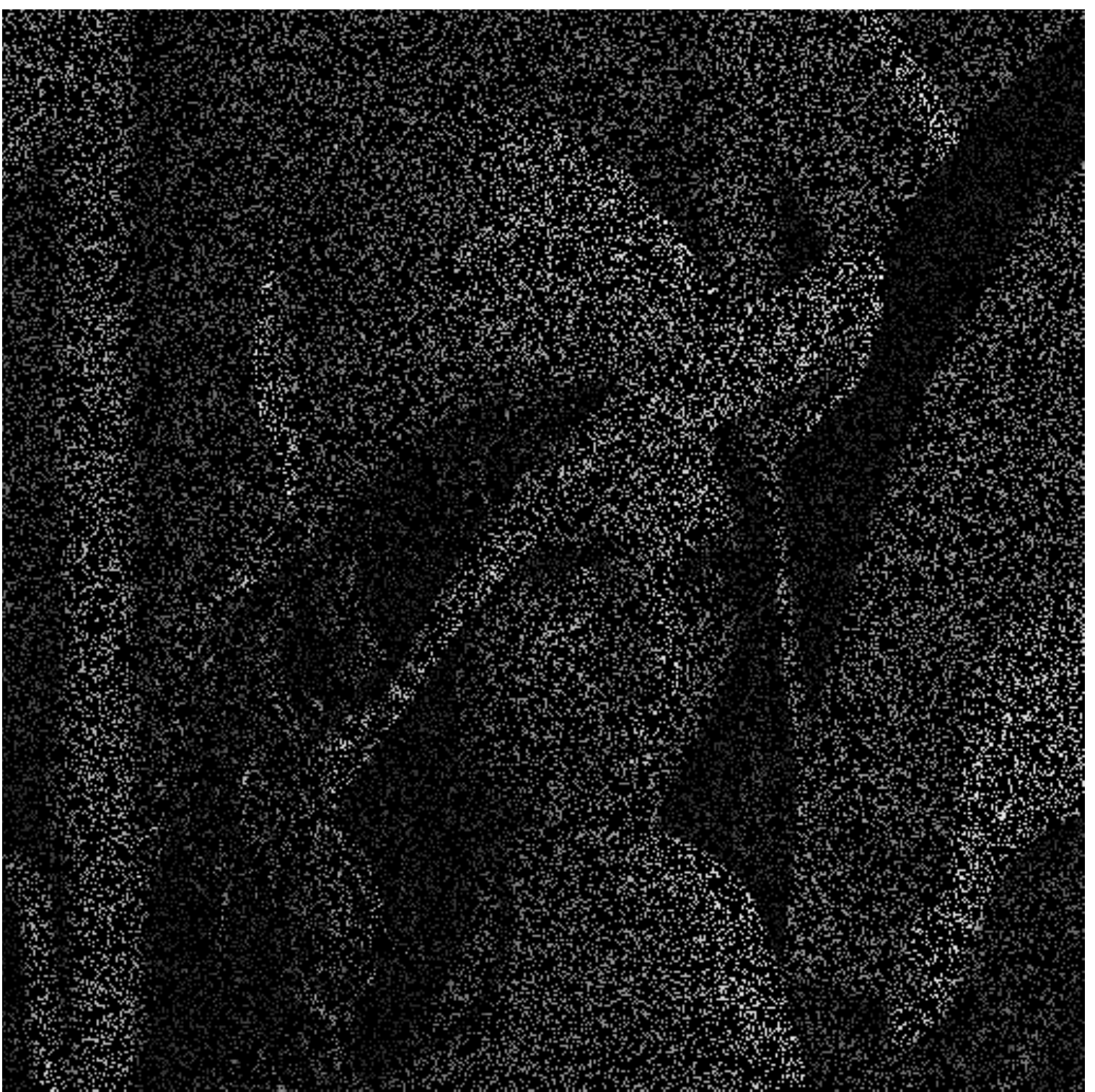} & \includegraphics[scale=0.205]{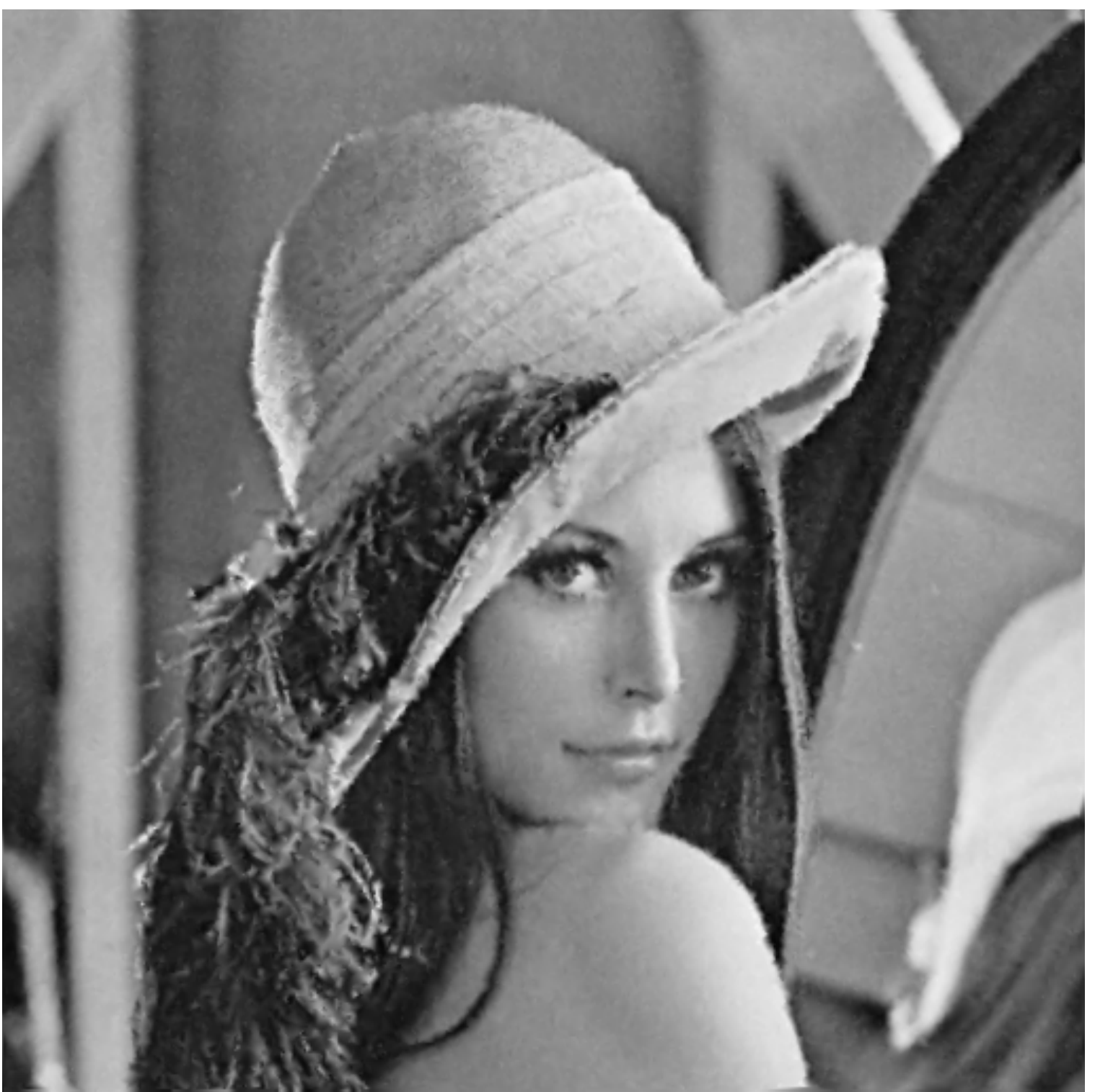}
& \includegraphics[scale=0.205]{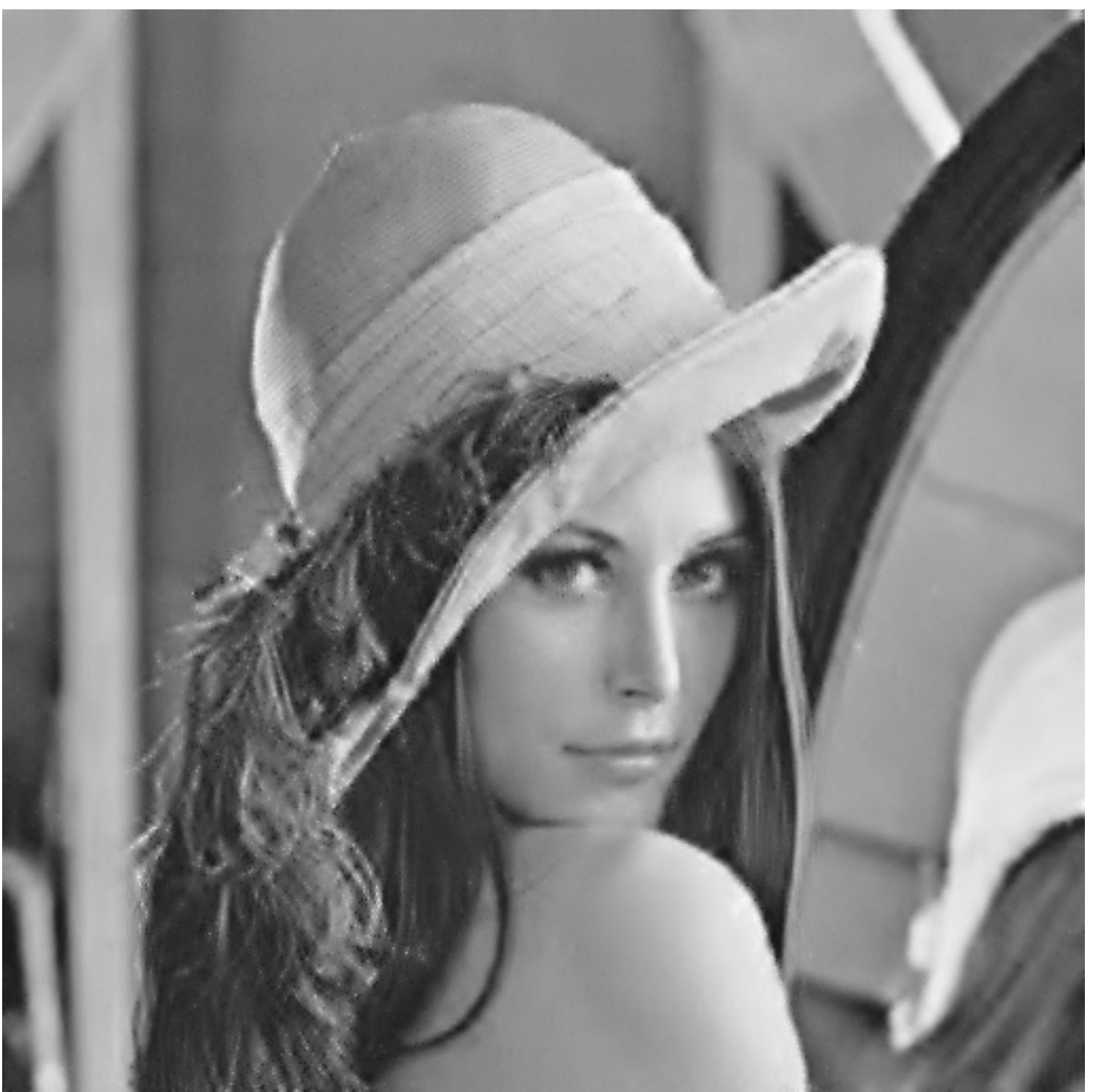} & \includegraphics[scale=0.205]{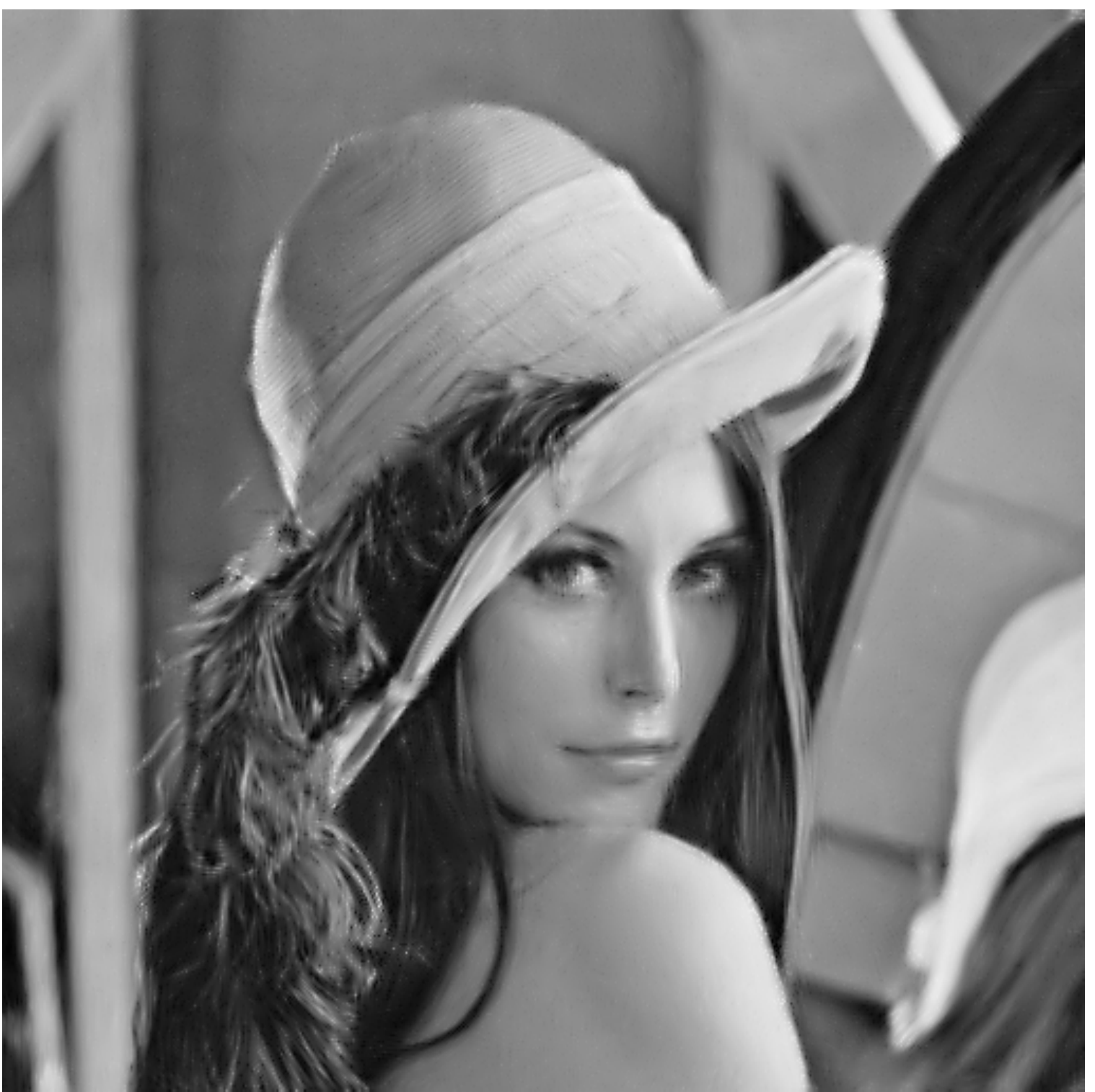} & \includegraphics[scale=0.205]{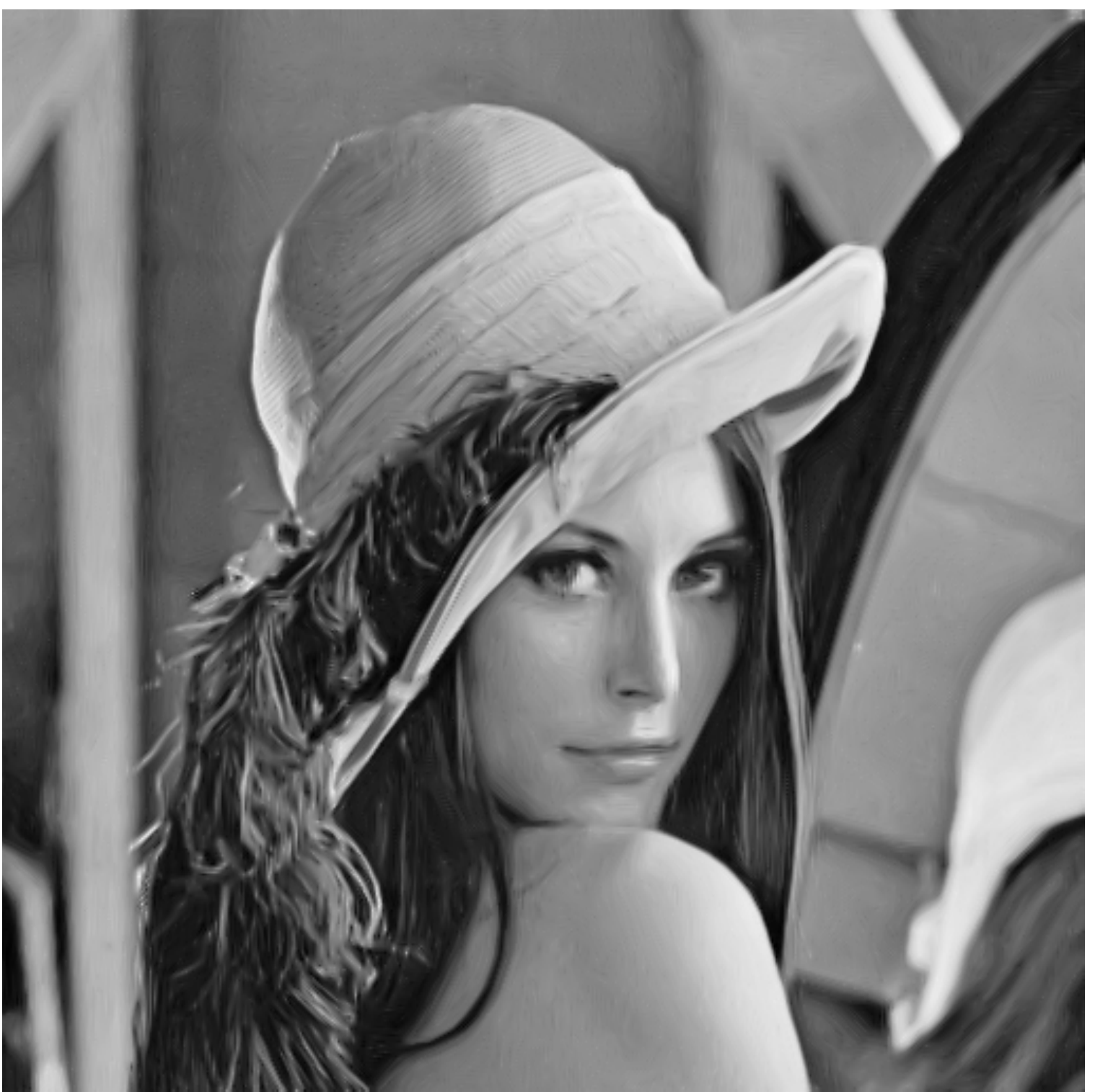}\\
{PSNR= 6.65 dB} & {PSNR=30.25 dB} & {PSNR=29.97 dB} & {PSNR=30.25 dB} & {PSNR=31.96 dB}\\
\includegraphics[scale=0.205]{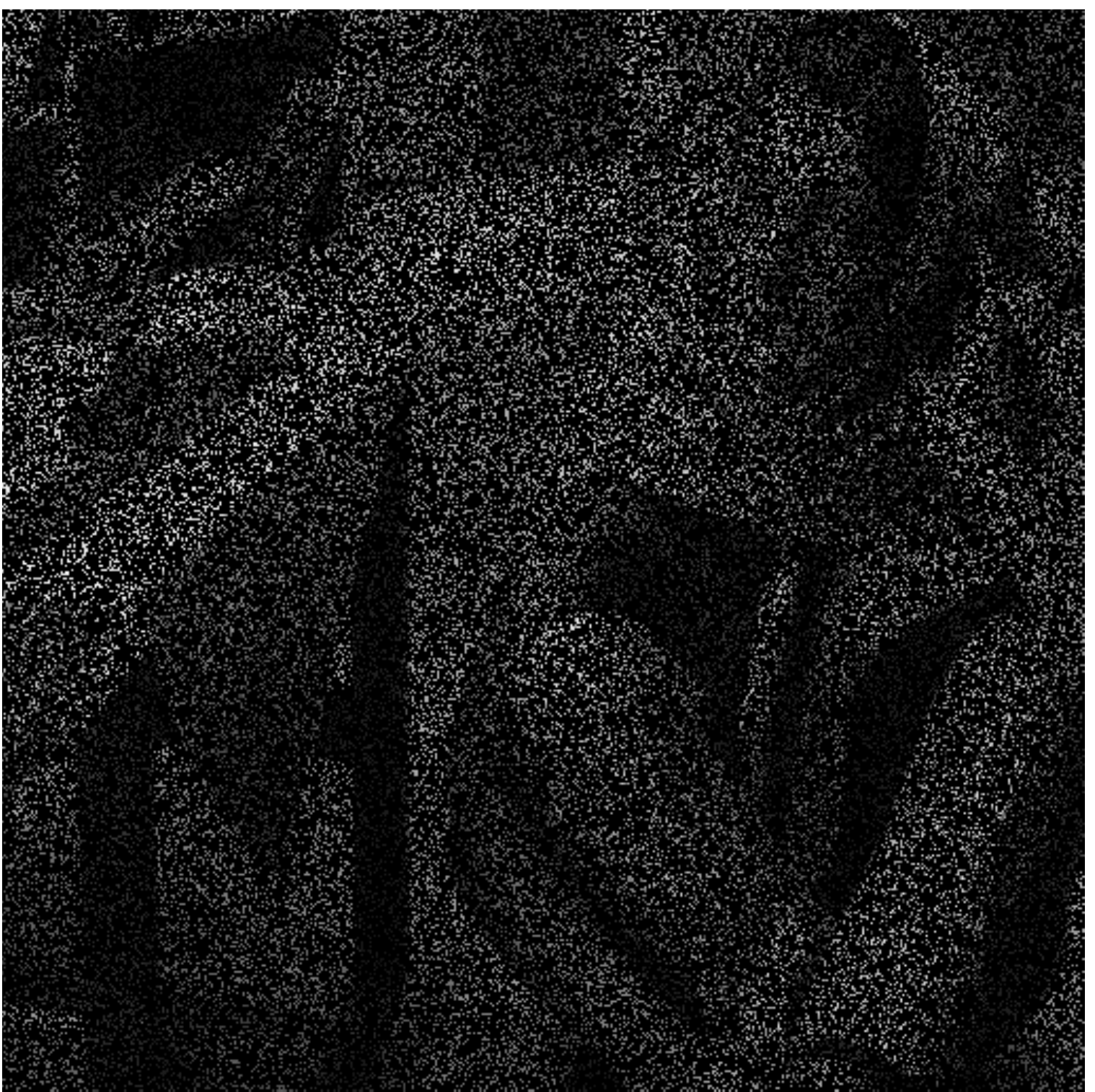} & \includegraphics[scale=0.205]{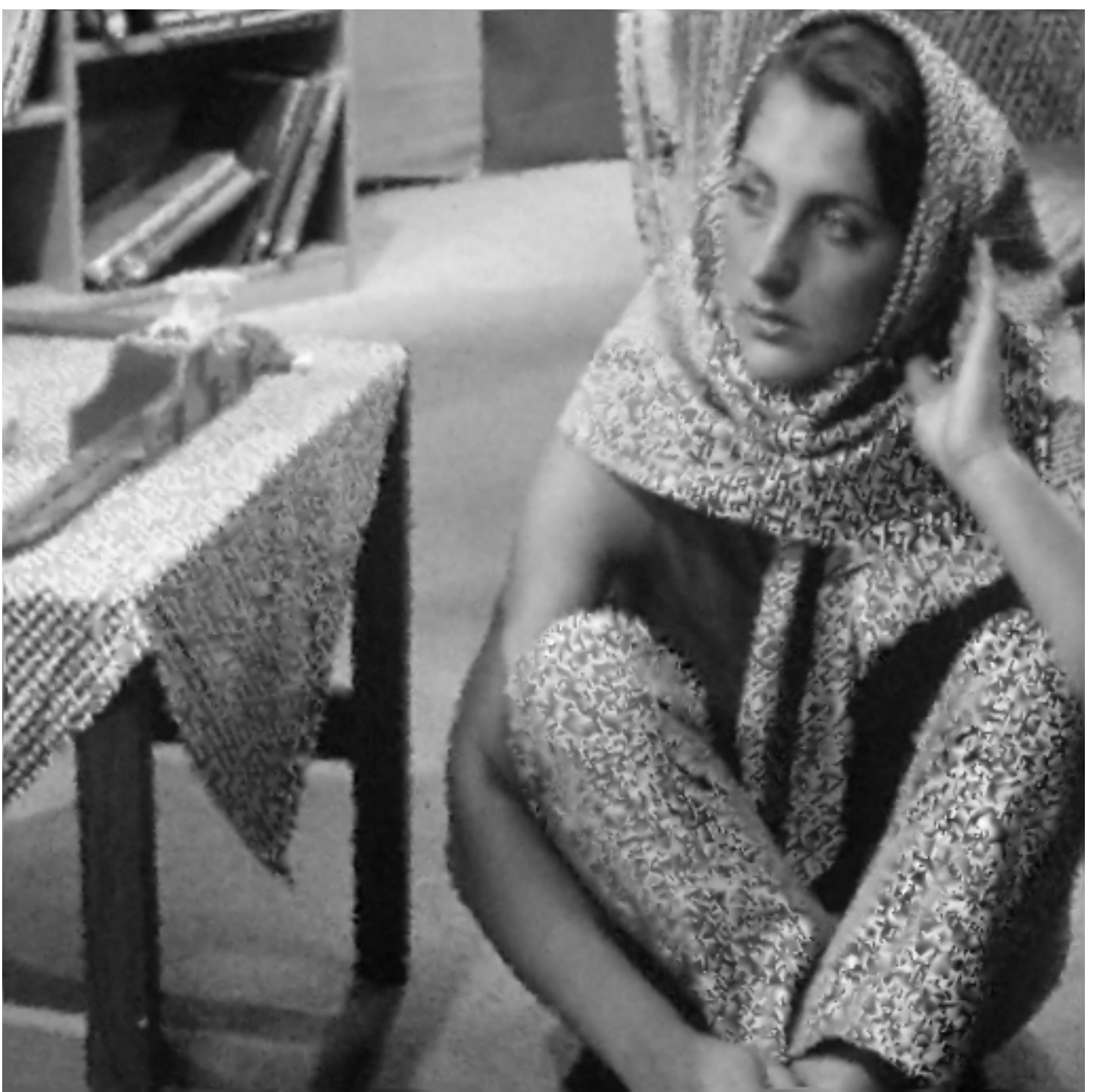}
& \includegraphics[scale=0.205]{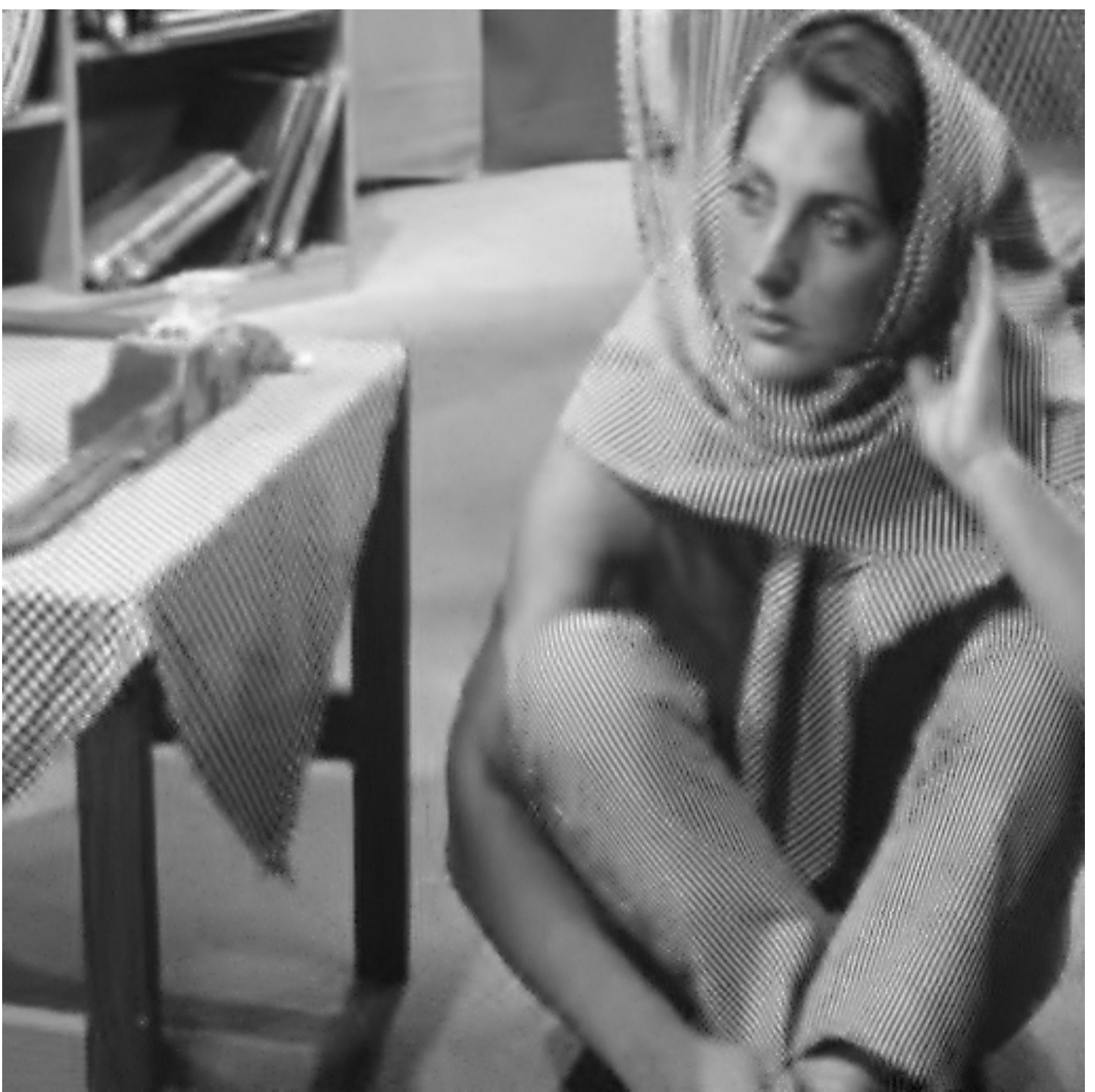} & \includegraphics[scale=0.205]{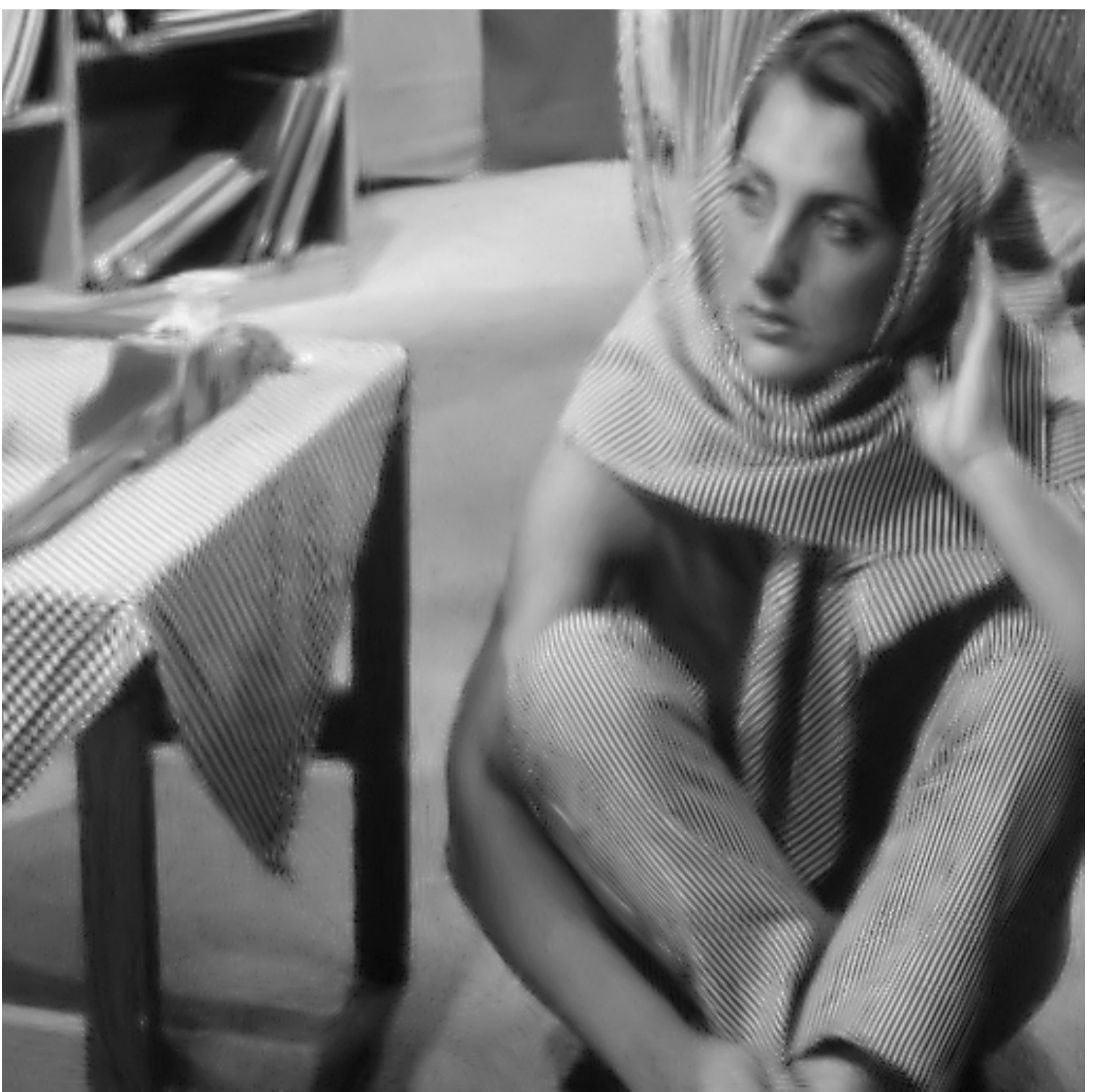} & \includegraphics[scale=0.205]{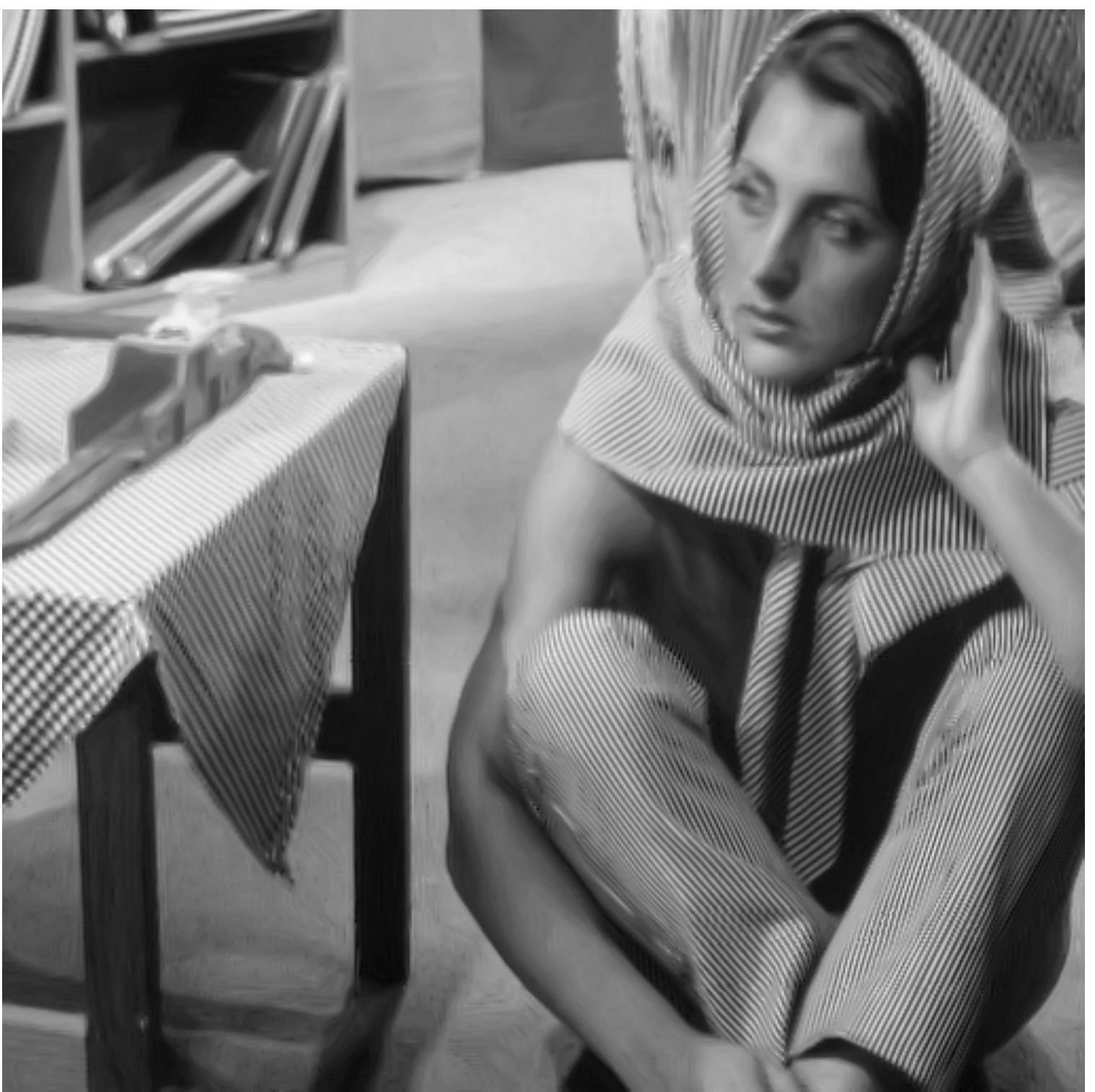}\\
{PSNR= 6.86 dB} & {PSNR=22.88 dB} & {PSNR=27.15 dB} & {PSNR=27.56 dB} & {PSNR=29.71 dB}\\
\includegraphics[scale=0.205]{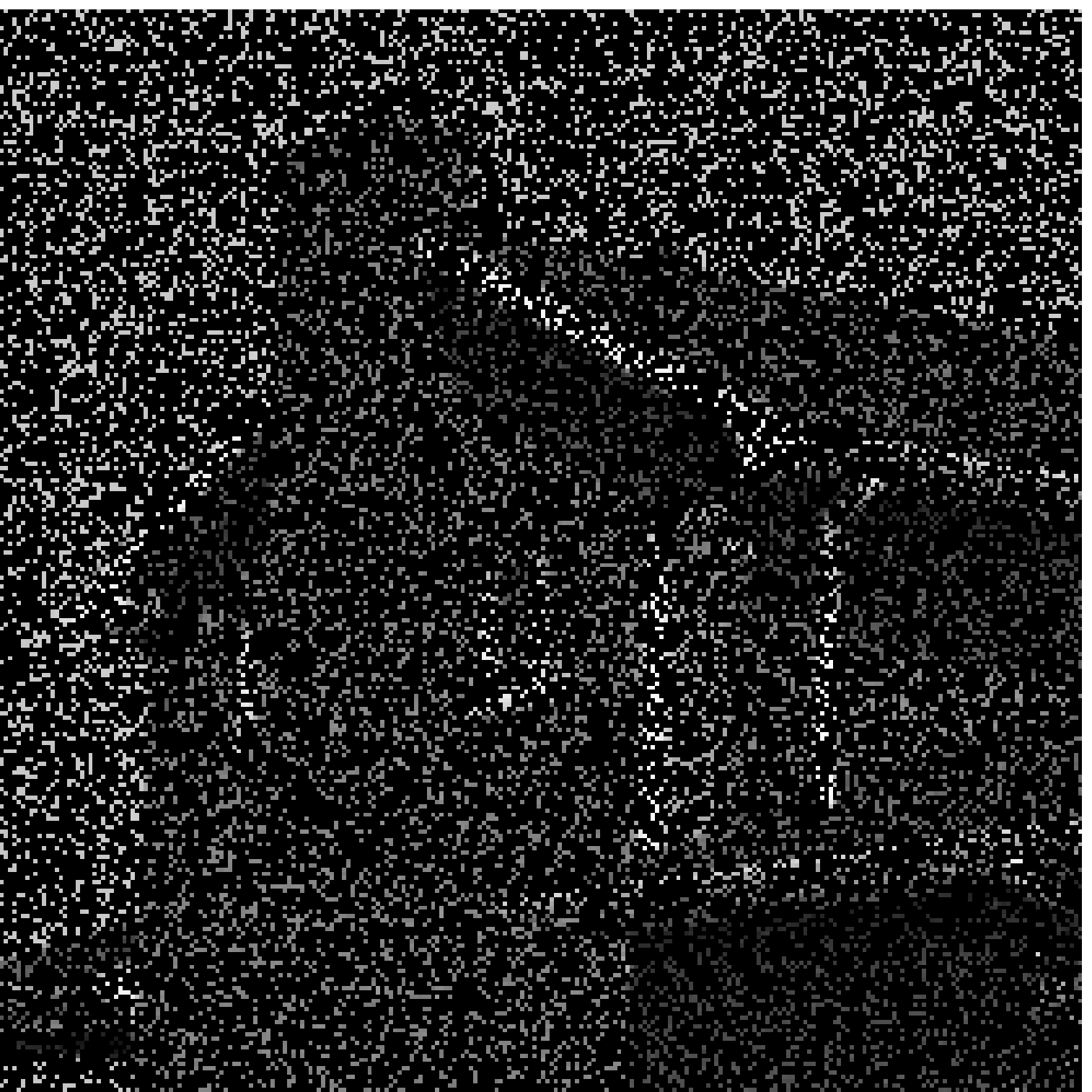} & \includegraphics[scale=0.205]{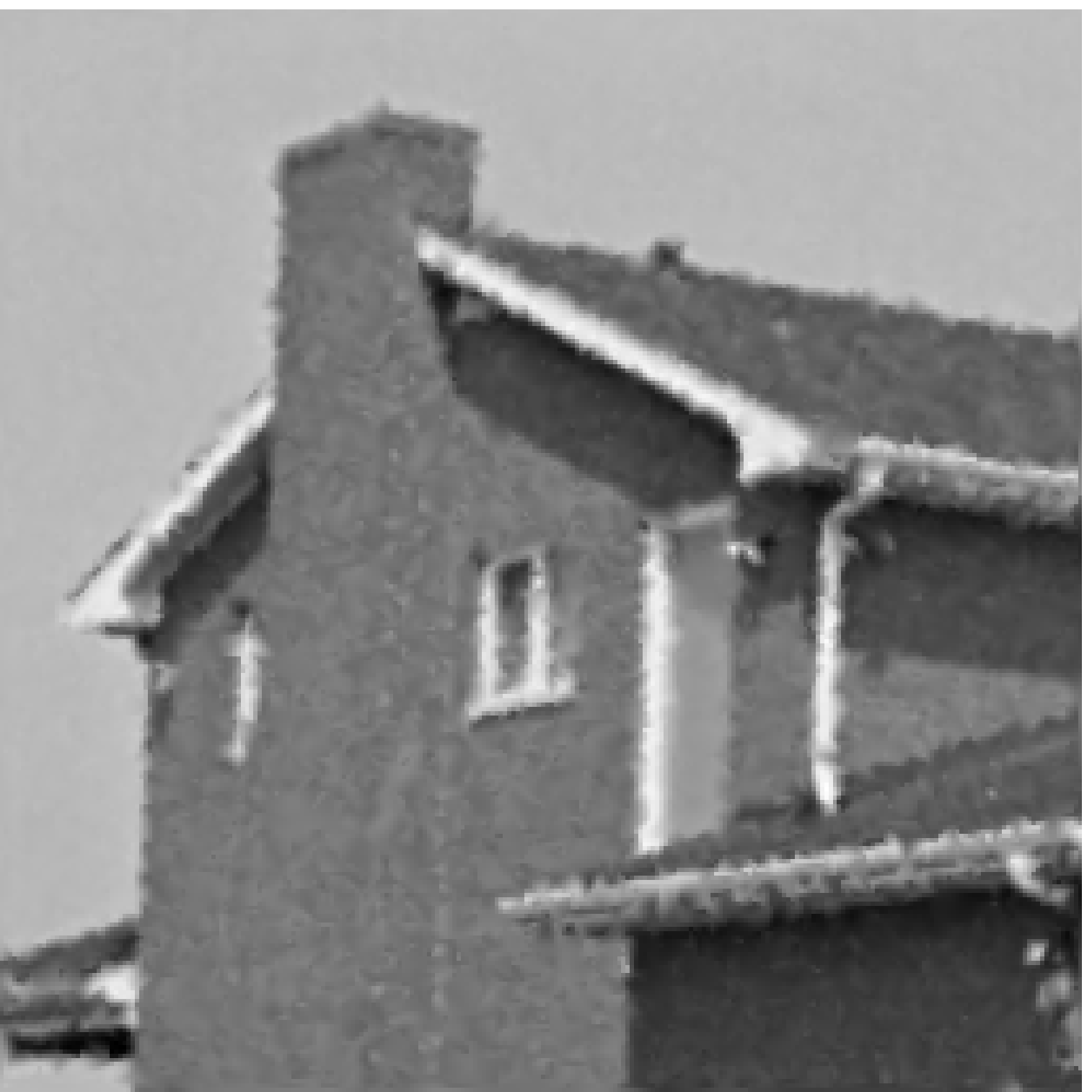}
& \includegraphics[scale=0.205]{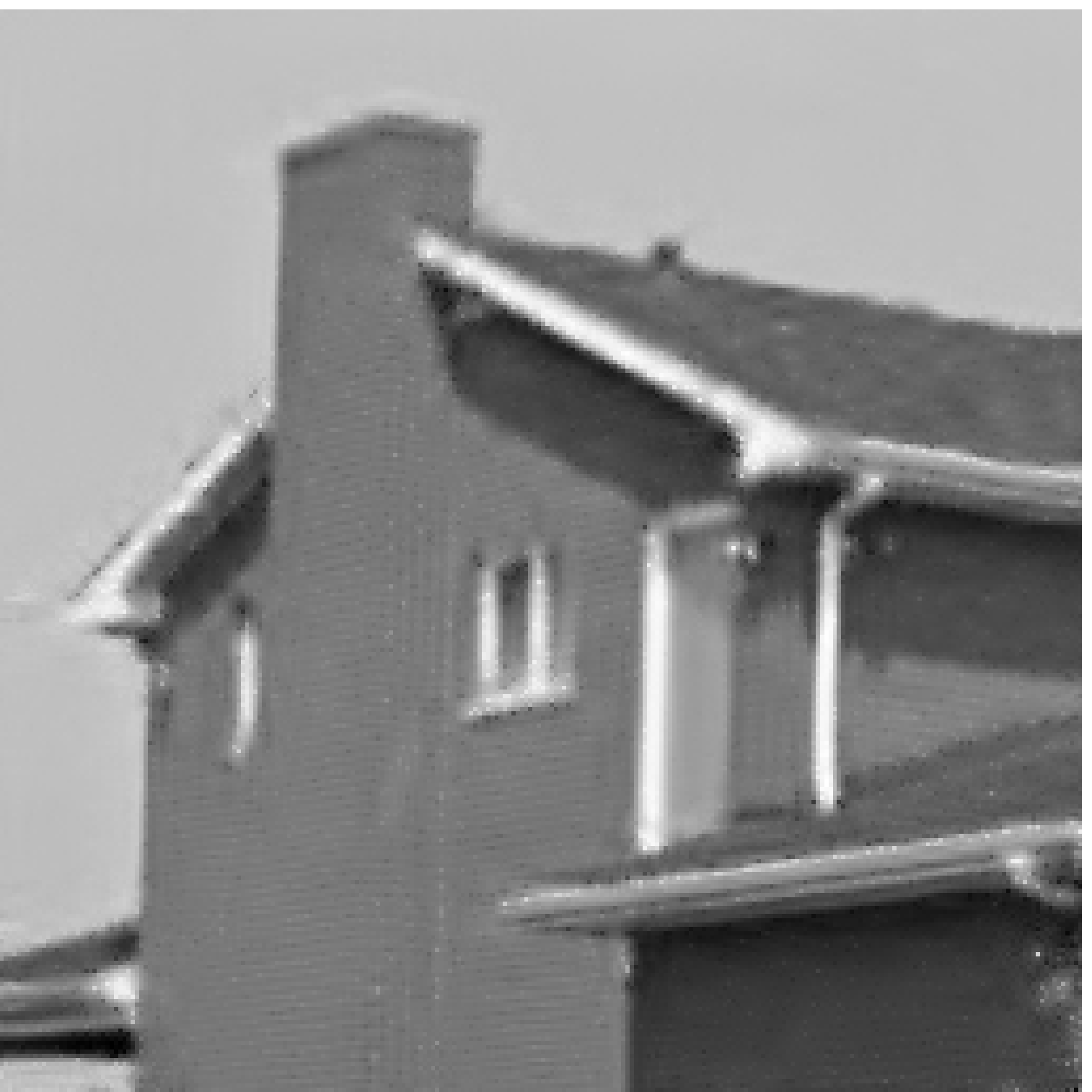} & \includegraphics[scale=0.205]{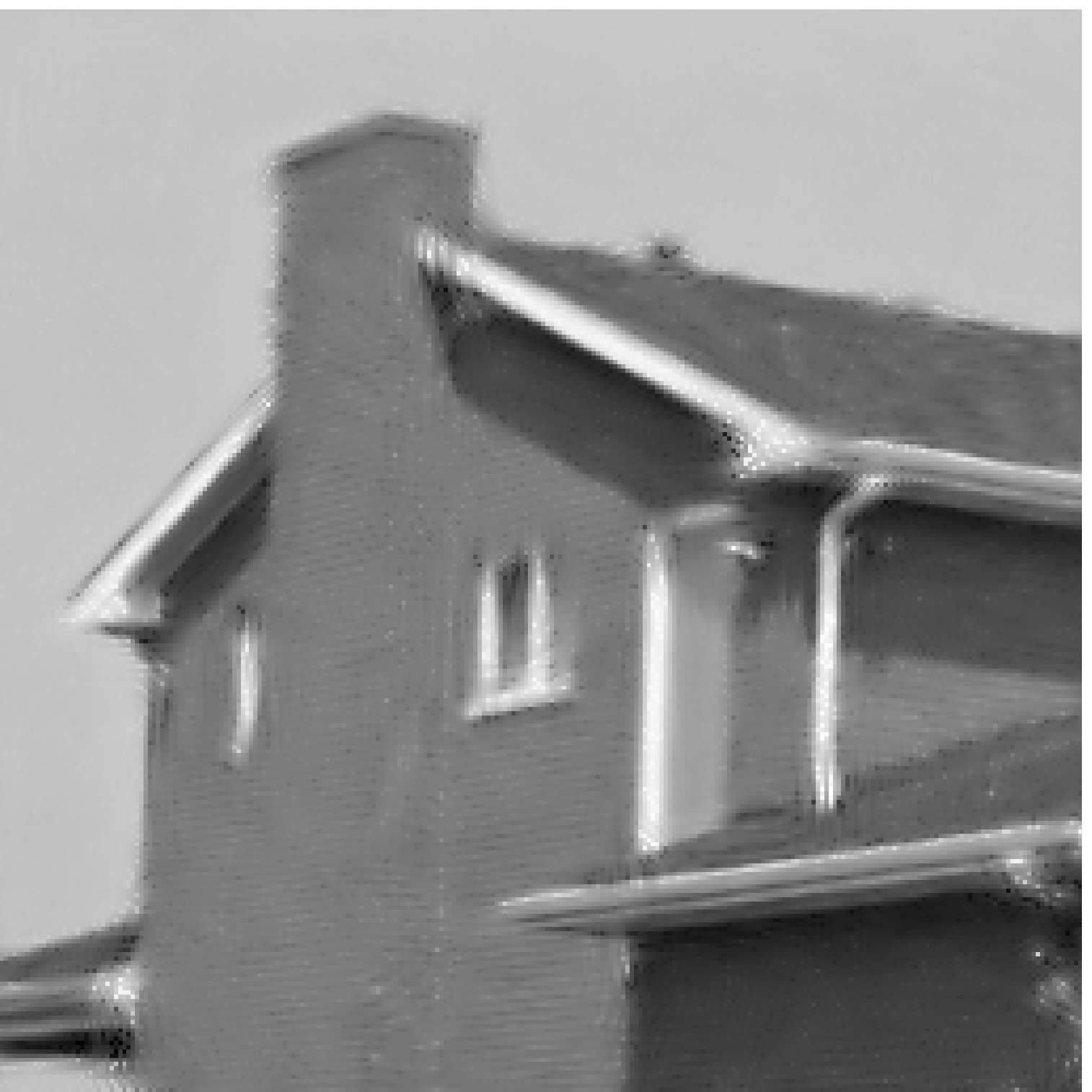} & \includegraphics[scale=0.205]{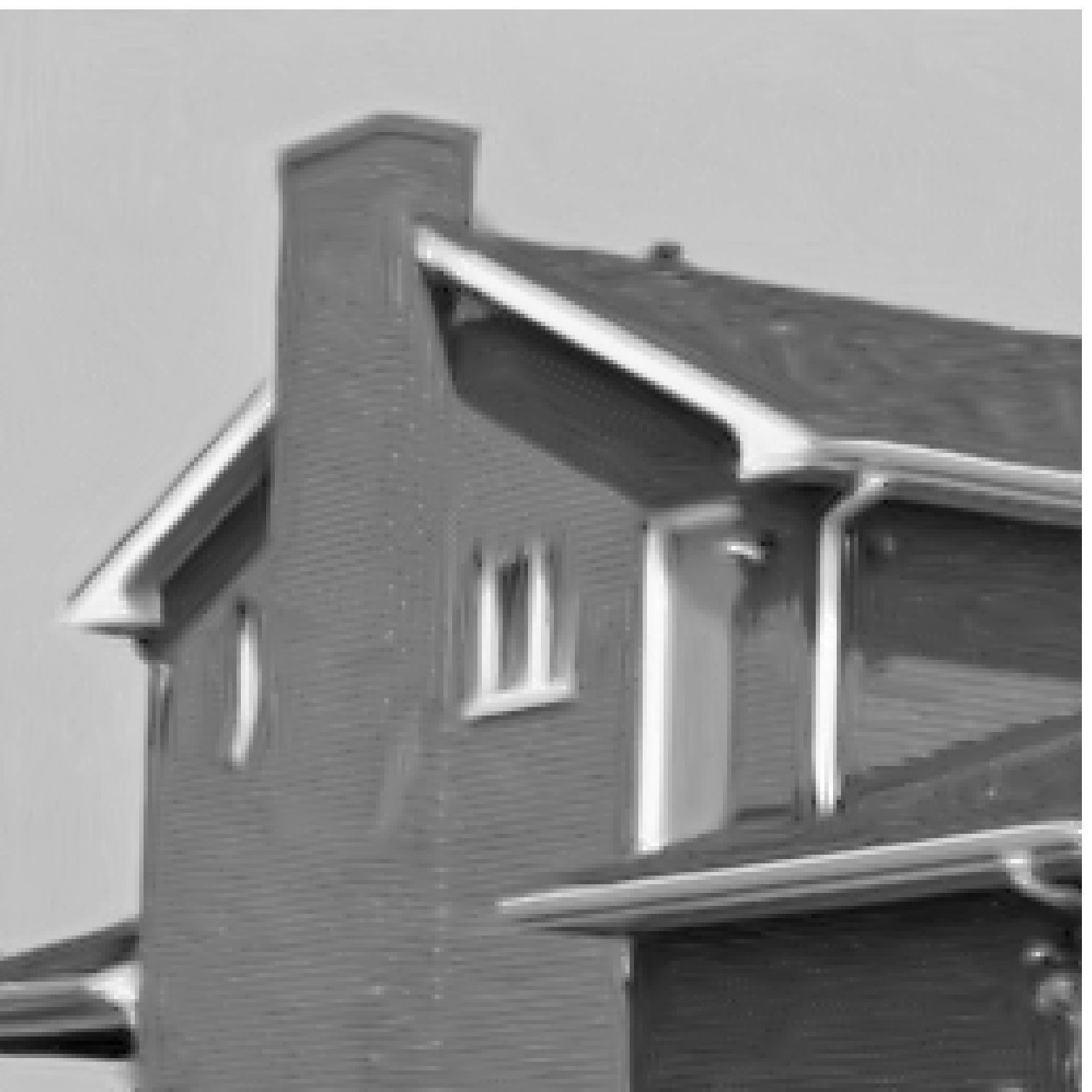}\\
{PSNR= 5.84 dB} & {PSNR= 29.21 dB} & {PSNR= 29.69 dB} & {PSNR= 29.03 dB} & {PSNR= 32.71 dB}\\
\end{tabular}
\caption{Inpainting results (PSNR) of corrupted versions of
the images Lena, Barbara and House with $80\%$ of their pixels missing (input PSNR=6.41 dB), obtained with different reconstruction methods:
: First column - corrupted images, Second column - triangle-based cubic interpolation,
Third column - overcomplete DCT dictionary, Fourth column - 1 iteration of the proposed scheme,
Fifth column - 3 iterations of the proposed scheme.}
\label{Figure: inpainted images}
\end{figure*}

We demonstrate the performance of our proposed scheme on corrupted versions of the images Lena, Barbara and House,
obtained by zeroing $80\%$ percents of their pixels, which are selected at random.
The parameters employed in each of the three iterations are shown in Table \ref{Table: inpainting parameters}.
In order to demonstrate the advantages of our method over simpler interpolation schemes
we compare our results to the ones obtained by the matlab function ``griddata''
which performs cubic interpolation of the missing pixels based on Delaunay triangulation \cite{yang1986finite},\cite{watson1992contouring}.
We also compare our results to the ones obtained using the algorithm described in chapter 15 of \cite{elad2010sparse},
which employs a patch-based sparse representation reconstruction algorithm with a DCT overcomplete dictionary to recover the image patches. We use a patch size of $16\times16$ pixels in order to improve the results this method produces.
We note that we do not employ the K-SVD based algorithm which was also described in this chapter,
as our experiments showed that it produces comparable or only slightly better results than the redundant DCT dictionary,
at a higher computational cost.
The PSNR values of the results obtained with the different algorithms are shown in Table \ref{Table: inpainting_PSNR}.
Fig.\ref{Figure: inpainted images} shows the corrupted and the reconstructed images, with the corresponding PSNR values,
obtained using triangle-based cubic interpolation,
overcomplete DCT dictionary, 1 and  3 iterations of the proposed scheme.
First, it can be seen that the second and third iterations greatly improve the results of our proposed algorithm.
It can also be seen that the results obtained with three iterations of our proposed scheme are much better than those obtained with the two other methods.

\begin{table}[t]
\centering
\caption{Parameters used in the inpainting experiments.\newline}
\begin{tabular}{|c|c|c|c|c|}\hline
Iteration & $K$ & $\sqrt{n}$ & $B$ & $\epsilon$\\\hline
1 & 10 & 16 & 9 & $10^2$\\\hline
2 & 10 & 8 & 43 & $10^4$\\\hline
3 & 10 & 5 & 55 & $10^8$\\\hline
\end{tabular}
\label{Table: inpainting parameters}
\end{table}

\begin{table}[t]
\centering
\caption{Inpainting results (PSNR in dB) of corrupted versions of
the images Lena, Barbara and House with 80 percents of their pixels missing,
obtained using triangle-based cubic interpolation (tri. cubic), overcomplete DCT dictionary, 1 and  3 iterations of the proposed scheme. For each image the best result is highlighted.\newline}
\begin{tabular}{|c|c|c|}\hline
Image & Method & PSNR\\\hline
\multirow{5}{*}{Lena} & tri. cubic & 30.25 \\\cline{2-3}
& DCT & 29.97 \\\cline{2-3}
& proposed (1 iter.) & 30.25 \\\cline{2-3}
& proposed (2 iter.) & 31.8 \\\cline{2-3}
& proposed (3 iter.) & $\mathbf{31.96}$ \\\noalign{\hrule height 1pt}
\multirow{5}{*}{Barbara} & tri. cubic & 22.88 \\\cline{2-3}
& DCT & 27.15 \\\cline{2-3}
& proposed (1 iter.) & 27.56 \\\cline{2-3}
& proposed (2 iter.) & 29.34 \\\cline{2-3}
& proposed (3 iter.) & $\mathbf{29.71}$ \\\noalign{\hrule height 1pt}
\multirow{5}{*}{House} & tri. cubic & 29.21 \\\cline{2-3}
& DCT & 29.69 \\\cline{2-3}
& proposed (1 iter.) & 29.03 \\\cline{2-3}
& proposed (2 iter.) & 32.1 \\\cline{2-3}
& proposed (3 iter.) & $\mathbf{32.71}$ \\\hline
\end{tabular}
\label{Table: inpainting_PSNR}
\end{table}

\subsection{Computational Complexity}

We next evaluate the computational complexity of a single iteration of the two image processing algorithms described above.
We note that for the image denoising scheme,
we assume that the filters training has been done beforehand, and exclude it from our calculations.
First, building the matrix $\mathbf{X}$ which contains the image patches requires $O(nN)$ operations.
We assume that when the nearest-neighbor search described above is used with a search window of size $B\times B$, most of the patches do not require to calculate distances outside this neighborhood.
Therefore, as calculating each of the distance measures (\ref{Euclidean Distance}) and (\ref{Inpainting Distance}) requires $O(n)$ operations, the number of operations required to calculate a single reordering matrix $\mathbf{P}_k$ can be bounded by $O(NB^2n)$.
Next, applying the matrices $\mathbf{P}_k$ and $\mathbf{P}_k^{-1}$ to the $n$ subimages $\mathbf{z}_j$ require $O(nN)$ operations,
and so does applying either one of the operators $H$ described above to the $n$ subimages $\mathbf{z}_j^p$.
Finally , constructing an estimate image by averaging the pixel values obtained with the different subimages also requires $O(nN)$ operations, and averaging the estimates obtained with the different matrices $\mathbf{P}_k$ requires $O(KN)$ operations.
Therefore when $K$ permutation matrices are employed, the total complexity is
\begin{align}
 O((n+K)N)+K\left[O(NB^2n)+O(nN)\right]=O(NKB^2n)
\end{align}
operations, which means that, as might be expected,
the overall complexity is dominated by the creation of the permutation matrices.
For a typical case in our experiments, $N=512^2$, $K=10$, $n=64$ and $B=111$,
and the above amounts to $1.86\cdot10^{10}$ operations.
We note that while in our experiments we employed exact exhaustive search,
approximate nearest neighbor algorithms may be used to alleviate the computational burden.

\section{Conclusions}

We have proposed a new image processing scheme which is based on smooth \ac{1D} ordering of the pixels in the given image.
We have shown that using a carefully designed permutation matrices and simple and intuitive 1D operations such as linear filtering and interpolation,
the proposed scheme can be used for image denoising and inpainting,
where it achieves high quality results.

There are several research directions to extend this work that we are currently considering.
The first is to make use of the distances between the patches not only to find the ordering matrices,
but also in the reconstruction process of the subimages.
These distances carry additional information which might improve the obtained results.
A different direction is to develop new image processing algorithms which involve optimization problems in which the 1D image reorderings act as regularizers.
These may both improve the image denoising and inpainting results, and allow to tackle other applications such as image deblurring, where the operator $\mathbf{M}$ is no longer restricted to be point-wise local.
Additionally, the proposed image denoising scheme may be improved by dividing the patches to more than two types,
and treating each type differently.
Finally, we note that in our work we have not exhausted the potential of the proposed algorithms,
and the choice of different parameters (e.g., $B,\epsilon$) for each set of patches may also improve the produced results.

\bibliographystyle{IEEEtran}
\bibliography{IEEEabrv,1Dreordering}

\end{document}